\documentclass[10pt,twocolumn,letterpaper]{article}

\usepackage{iccv}
\usepackage{times}
\usepackage{epsfig}
\usepackage{graphicx}
\usepackage{amsmath}
\usepackage{amssymb}

\usepackage{subfig}
\usepackage{booktabs}
\usepackage{bm}
\usepackage{comment}
\usepackage{color,colortbl}
\usepackage{mathtools}
\usepackage{xfrac}
\usepackage{multirow}
\usepackage{nicefrac}
\usepackage{caption}
\usepackage{stmaryrd}
\usepackage{lipsum}  
\usepackage{csquotes}  
\usepackage{verse}
\usepackage{soul}
\usepackage{authblk}

\definecolor{tabbestcolor}{rgb}{0.785, 0.851, 0.969}
\def \best {\cellcolor{tabbestcolor!85}}
\def \sbest {\cellcolor{tabbestcolor!30}}

\usepackage{pifont}

\newcommand{\Acronym}[0]{DeLiRa\xspace} 
\newcommand{\Section}[0]{Sec.\xspace} 
\newcommand{\Equation}[0]{Eq.\xspace} 
\newcommand{\Figure}[0]{Fig.\xspace} 
\newcommand{\Table}[0]{Tab.\xspace} 

\definecolor{Gray}{gray}{0.925}
\definecolor{White}{gray}{1.0}

\makeatletter
\newcommand\notsotiny{\@setfontsize\notsotiny{5.7}{7}}
\makeatother

\usepackage[pagebackref=true,colorlinks,bookmarks=false]{hyperref}

\iccvfinalcopy 

\def\iccvPaperID{5350} 

\begin{document}

\title{
DeLiRa: Self-Supervised Depth, Light, and Radiance Fields
}

\author[ ]{
Vitor Guizilini$^1$ 
\quad\quad 
Igor Vasiljevic$^1$ 
\quad\quad 
Jiading Fang$^2$
\quad\quad 
Rares Ambrus$^1$
\quad\quad 
Sergey Zakharov$^1$
\quad\quad 
Vincent Sitzmann$^3$
\quad\quad 
Adrien Gaidon$^1$
}

\affil[1]{\small Toyota Research Institute (TRI), Los Altos, CA}
\affil[2]{\small Toyota Technological Institute of Chicago (TTIC), Chicago, IL}
\affil[3]{\small Massachusetts Institute of Technology (MIT), Cambridge, MA}


\twocolumn[{%
\renewcommand\twocolumn[1][]{#1}%
\maketitle
\vspace{-8mm}
\begin{center}
    \centering
    \captionsetup{type=figure}
    \includegraphics[height=3.6cm]{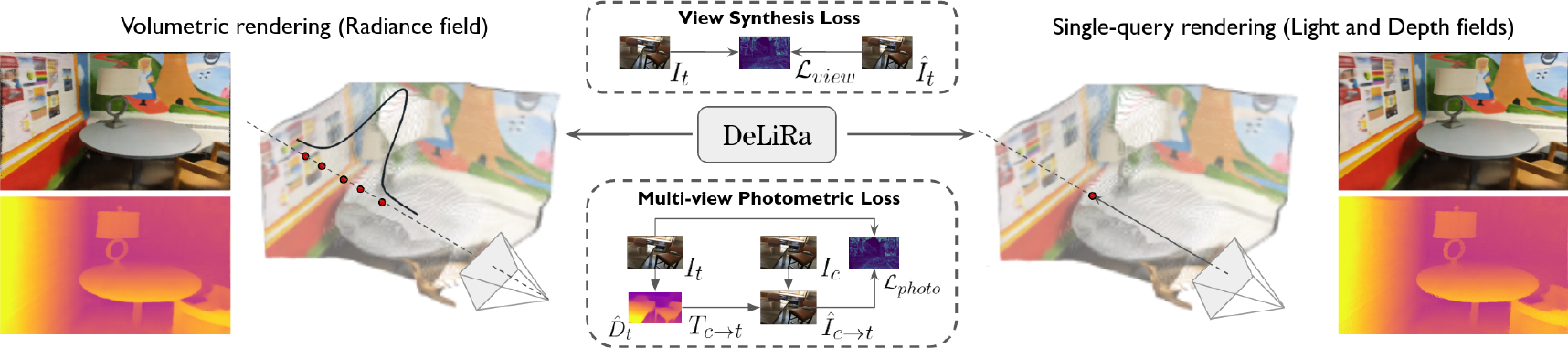}    
    \captionof{figure}{
    \textbf{\Acronym augments volumetric view synthesis with the multi-view photometric objective}, as a regularizer to improve novel view and depth synthesis in the limited viewpoint setting. We use this implicit representation to jointly learn depth, light, and radiance fields from a shared latent space in a synergistic way.
    }
    \label{fig:teaser}
\end{center}
    }]

\maketitle
\begin{abstract}
Differentiable volumetric rendering is a powerful paradigm for 3D reconstruction and novel view synthesis.
However, standard volume rendering approaches struggle with degenerate geometries in the case of limited viewpoint diversity, a common scenario in robotics applications.
In this work, we propose to use the multi-view photometric objective from the self-supervised depth estimation literature as a geometric regularizer for volumetric rendering, significantly improving novel view synthesis without requiring additional information.
Building upon this insight, we explore the explicit modeling of scene geometry using a generalist Transformer, jointly learning a radiance field as well as depth and light fields with a set of shared latent codes.
We demonstrate that sharing geometric information across tasks is mutually beneficial, leading to improvements over single-task learning without an increase in network complexity.
Our \Acronym architecture achieves state-of-the-art results on the ScanNet benchmark, enabling high quality volumetric rendering as well as real-time novel view and depth synthesis in the limited viewpoint diversity setting.
Our project page is \href{https://sites.google.com/view/tri-delira}{https://sites.google.com/view/tri-delira}.
\end{abstract}

\section{Introduction}

Inferring 3D geometry from 2D images is a cornerstone capability in computer vision and computer graphics. In recent years, the state of the art has significantly advanced due to the development of  neural fields~\cite{xie2022neural}, which parameterize continuous functions in 3D space using neural networks, and differentiable rendering~\cite{tewari2022advances,mildenhall2020nerf,sitzmann2019scene}, which enables learning these functions directly from images.
However, recovering 3D geometry from 2D information is an ill-posed problem: there is an inherent ambiguity of shape and radiance (i.e. the \emph{shape-radiance ambiguity}~\cite{nerf++}). These representations thus require a large number of diverse camera viewpoints in order to converge to the correct geometry.
Alternatively, methods that explicitly leverage geometric priors at training time, via the self-supervised multi-view photometric objective, have achieved great success for tasks such as depth~\cite{monodepth2,packnet,watson2021temporal,tri-depthformer}, ego-motion~\cite{tang2020neural,tang2020kp3d}, camera geometry~\cite{vasiljevic2020neural,tri-self_calibration,gordon2019depth}, optical flow~\cite{tri_draft_ral22}, and scene flow~\cite{tri_draft_ral22,selfsceneflow}.

In this work, we combine these two paradigms and introduce the multi-view photometric loss as a complement to the view synthesis objective. 
Specifically, we use depth inferred via volumetric rendering to warp images, with the photometric consistency between synthesized and original images serving as a self-supervisory regularizer to scene structure.
We show through experiments that this explicit regularization facilitates the recovery of accurate geometry in the case of low viewpoint diversity, without requiring additional data.
Because the multi-view photometric objective is unable to model view-dependent effects (since it assumes a Lambertian scene), we propose an attenuation schedule that gradually removes it from the optimization, and show that our learned scene geometry is stable, leading to further improvements in view and depth synthesis. 

We take advantage of this accurate learned geometry and propose \textbf{\Acronym}, an auto-decoder architecture inspired by~\cite{jaegle2021perceiverio} that jointly estimates \textbf{De}pth~\cite{tri-define},  \textbf{Li}ght~\cite{sitzmann2021light}, and \textbf{Ra}diance~\cite{mildenhall2020nerf} fields.
We maintain a shared latent representation across task-specific decoders, and  show that this increases the expressiveness of learned features and is beneficial for all considered tasks, improving performance over single-task networks without additional complexity. 
Furthermore, we explore other synergies between these representations: volumetric predictions are used as pseudo-labels for the depth and light fields, improving viewpoint generalization; and depth field predictions are used as guidance for volumetric sampling, significantly improving efficiency without sacrificing performance. 

To summarize, our contributions are as follows. In our first contribution, we show that the \textbf{multi-view photometric objective is an effective regularization tool for volumetric rendering}, as a way to mitigate the shape-radiance ambiguity.
To further leverage this geometrically-consistent implicit representation, in our second contribution we propose a \textbf{novel architecture for the joint learning of depth, light, and radiance fields}, decoded from a set of shared latent codes. 
We show that jointly modeling these three fields leads to improvements over single-task networks, without requiring additional complexity in the form of regularization or image-space priors.
As a result, \textbf{our proposed method achieves state-of-the-art view synthesis and depth estimation results on the ScanNet benchmark}, outperforming methods that require explicit supervision from ground truth or pre-trained networks.
\section{Related Work}

\subsection{Implicit Representations for View Synthesis}

Our method falls in the category of auto-decoder architectures for neural rendering~\cite{park2019deepsdf,xie2022neural}, which directly optimize a latent code. %
Building on top of DeepSDF~\cite{park2019deepsdf}, SRN~\cite{sitzmann2019scene} adds a differentiable ray marching algorithm to regress color, enabling training from a set of posed images.
The vastly popular NeRF~\cite{mildenhall2020nerf} family regresses color and density, using volumetric rendering to achieve state-of-the-art free view synthesis.
CodeNeRF\cite{jang2021codenerf} learns instead the variation of object shapes and textures, and does not require knowledge of camera poses at test time.

Despite recent improvements, efficiency remains one of the main drawbacks of volumetric approaches, since rendering each pixel requires many network calls.
To alleviate this, some methods have proposed better sampling strategies \cite{hu2022efficientnerf, wang2021neus}. 
This is usually achieved using depth priors,
either from other sensors~\cite{rematas2021urban}, sparse COLMAP~\cite{colmap} predictions~\cite{dsnerf}, or pre-trained depth networks~\cite{nerfingmvs,depthpriorsnerf,donerf}.
Other methods have moved away from volumetric rendering altogether and instead generate predictions with a single forward pass~\cite{sitzmann2021light,tri-define,wang2022r2l}. 
While much more efficient, these methods require substantial viewpoint diversity to achieve the multi-view consistency inherent to volumetric rendering. 
Light field networks~\cite{sitzmann2021light} map an oriented ray directly to color, relying on generalization to learn a multi-view consistency prior. 
DeFiNe~\cite{tri-define} learns \textit{depth} field networks, using ground truth to generate virtual views via explicit projection.
R2L~\cite{wang2022r2l} uses a pre-trained volumetric model, that is distilled into a residual light field network.

Our proposed method combines these two directions 
into a single framework. 
Differently from DeFiNe and R2L, it does not require ground truth depth maps or a pre-trained volumetric model for distillation. 
Instead, we propose to \textit{jointly} learn self-supervised depth, light, and radiance fields, decoded from the same latent space.  
A radiance decoder generates volumetric predictions that serve as additional multi-view supervision for light and depth field decoders. 
At the same time, predictions from the depth field decoder serve as priors to improve
sampling for volumetric rendering, decreasing the amount of required network calls.

\begin{figure*}[t!]
\centering
\subfloat[Geometric embeddings, including camera center $\mathbf{t}_t$, viewing rays $\mathbf{r}_{ij}$, and sampled 3D points $\mathbf{x}_k$ from each viewing ray.]{
\includegraphics[width=0.45\textwidth]{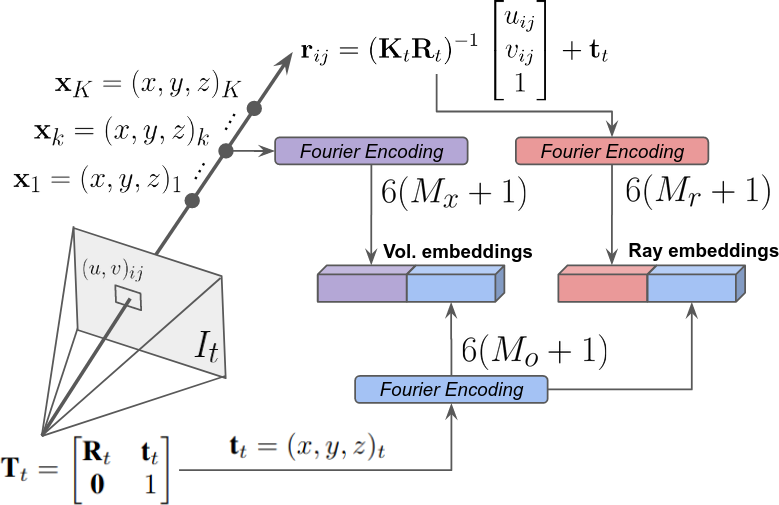}
\label{fig:geometric_queries}
}
\quad
\subfloat[
Multi-task depth, light, and radiance decoding. Geometric embeddings are used as queries $Q$, and cross-attended with keys $K$ and values $V$ from  $\mathcal{S}$.
]{
\raisebox{10mm}{\includegraphics[width=0.49\textwidth]{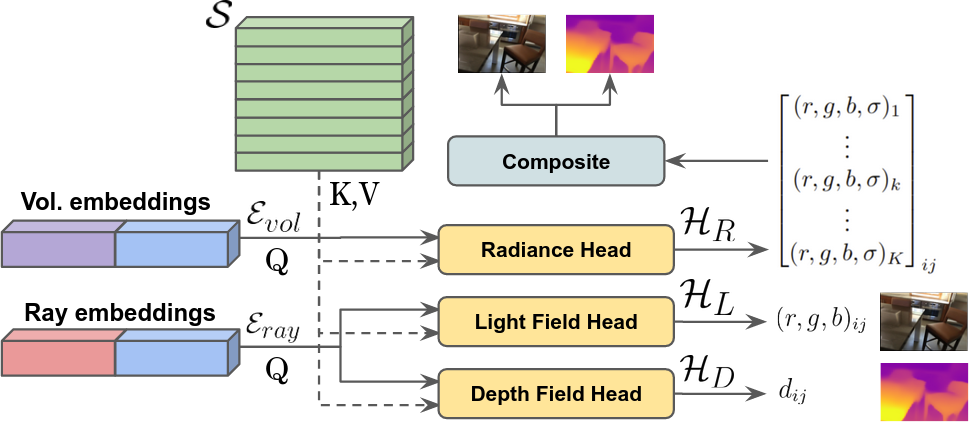}}
\label{fig:decoding}
}
\caption{\textbf{Diagram of our proposed \Acronym architecture}. In (a) we show how the various geometric embeddings are calculated from camera information, and in (b) we show depth, light, and radiance decoding from the same latent space $\mathcal{S}$.}
\label{fig:diagram}
\vspace{-4mm}
\end{figure*}


\subsection{Self-Supervised Depth Estimation}

The work of Godard \etal~\cite{godard2017unsupervised} introduced  self-supervision to the task of depth estimation, by framing it as a view synthesis problem:
given a target and context images, we can use predicted depth and relative transformation to warp information between viewpoints. 
By minimizing a photometric objective between target and synthesized images, depth and relative transformation are learned as proxy tasks. 
Further improvements to this original framework, in terms of losses~\cite{monodepth2,shu2020featdepth,tri-fsm}, camera  modeling~\cite{gordon2019depth,vasiljevic2020neural,klingner2020self,tri-self_calibration} and network architectures~\cite{packnet,tri-depthformer,simplified}, have led to performance comparable to or even better than supervised methods. 
We extend this self-supervised learning paradigm to the volumetric rendering setting, introducing it as an additional source of regularization to the single-frame view synthesis objective.  Our argument is that, by enforcing explicit multi-view photometric consistency in addition to implicit density-based volumetric rendering, we avoid degenerate geometries during the learning process.


\subsection{Structure Priors for Volumetric Rendering}

A few works have recently started exploring how to incorporate depth and structure priors in the volumetric rendering setting~\cite{donerf,dsnerf,nerfingmvs,depthpriorsnerf}. Most use structure-from-motion (e.g., COLMAP~\cite{colmap}) pointclouds, predicted jointly with camera poses, as ``free" supervision to regularize volumetric depth estimates. However, because these pointclouds are noisy and very sparse ($<0.1\%$ of valid projected pixels), substantial post-processing is required. 
RegNeRF~\cite{niemeyer2022regnerf} uses a normalizing-flow-based likelihood model over image patches to regularize predictions from unobserved views.
DS-NeRF~\cite{dsnerf} uses reprojection error as a measure of uncertainty, and minimizes the KL-divergence between volumetric and predicted depth.
NerfingMVS~\cite{nerfingmvs} trains a separate depth prediction network, used for depth-guided sampling.
DDP-NeRF~\cite{depthpriorsnerf} goes further and trains a separate depth \textit{completion} network, that takes SfM pointclouds as additional input to generate dense depth maps for supervision and sampling guidance. 
SinNeRF~\cite{xu2022sinnerf} uses a single RGB-D image to learn a radiance field.  Like our method, they use multi-view photometric warping as additional supervision, but crucially they rely on \textit{ground truth} depth, and thus the two objectives (volume rendering and warp-based view synthesis) are decoupled.
MonoSDF~\cite{yu2022monosdf} uses a pre-trained depth network to supervise SDF-based volume rendering, achieving impressive improvements in depth estimation, albeit at the expense of novel view synthesis performance.

Importantly, all these methods require additional data to train their separate networks, in order to generate the depth priors used for volumetric rendering. 
NerfingMVS uses a network pre-trained on $170$K samples from $4690$ COLMAP-annotated sequences~\cite{mannequin}. 
DDP-NeRF uses a network trained on $94$K RGB-D in-domain samples (i.e., from the same dataset used for evaluation). 
In these two methods (and several others~\cite{deng2021depth, yen2022nerf}), supervision comes from ``free" noisy COLMAP pointclouds. Drawing from the self-supervised depth estimation literature, we posit that geometric priors learned with a multi-view photometric objective are a stronger source of ``free" supervision.

\section{Methodology}

Our goal is to learn an implicit 3D representation from a collection of RGB images $\{I_i\}_{i=0}^{N-1}$. 
For each camera, we assume known intrinsics $\textbf{K}_i \in \mathbb{R}^{3\times 3}$ and extrinsics $\mathbf{T}_i \in \mathbb{R}^{4 \times 4}$, obtained as a pre-processing step~\cite{colmap}. 
Note that we assume neither ground-truth~\cite{donerf} nor pseudo-ground truth~\cite{donerf, nerfingmvs,dsnerf,depthpriorsnerf} depth supervision.

\subsection{Shared Latent Representation}

Our architecture for the joint learning of depth, light, and radiance fields (\Acronym) stores scene-specific information as a latent space $\mathcal{S} \in \mathbb{R}^{N_l \times C_l}$, composed of $N_l$ vectors with dimensionality $C_l$.
Cross-attention layers are used to decode queries, containing geometric embeddings (\Figure \ref{fig:geometric_queries}), into task-specific predictions. 
To self-supervise these predictions, we combine the view synthesis objective on RGB estimates and the multi-view photometric objective on depth estimates. 
We also explore other cross-task synergies, namely how volumetric predictions can be used to increase viewpoint diversity for light and depth field estimates, and how depth field predictions can serve as priors for volumetric importance sampling.
A diagram of \Acronym is shown in \Figure \ref{fig:decoding}, and below we describe each of its components.

\subsection{Geometric Embeddings}

We use geometric embeddings to process camera information, that serve as queries to decode from the latent space $\mathcal{S}$. 
Let $\textbf{u}_{ij} = (u,v)$ be an image coordinate in target camera $t$, with assumed known pinhole intrinsics $\mathbf{K}_t$, resolution $H \times W$, and extrinsics $\textbf{T}_t= \left[
\begin{smallmatrix}
\mathbf{R} & \textbf{t} \\
\textbf{0} & 1
\end{smallmatrix}\right]$
relative to camera $T_0$. 
Its origin $\mathbf{o}_t$ and direction $\textbf{r}_{ij}$ vectors are given by:
\begin{equation}
\small
\textbf{o}_t = - \mathbf{R}_t \mathbf{t}_t 
\quad , \quad 
\textbf{r}_{ij} = \big(\mathbf{K}_t \mathbf{R}_t \big)^{-1}  
\left[u_{ij},v_{ij},1\right]^T 
\end{equation}
Note that direction vectors are normalized to unitary values $\bar{\textbf{r}}_{ij} = \frac{\textbf{r}_{ij}}{||\textbf{r}_{ij}||}$ before further processing. 
For volumetric rendering, we sample $K$ times along the viewing ray to generate 3D points $\textbf{x}_k=(x,y,z)$ given depth values $z_k$:
\begin{equation}
\small
\textbf{x}_{ij}^k = \textbf{o}_t + z_k\bar{\textbf{r}}_{ij}
\end{equation}
In practice, $z_k$ values are linearly sampled between a $[d_{min},d_{max}]$ range.
These vectors, $\textbf{o}_t$, $\bar{\textbf{r}}_{ij}$ and $\textbf{x}_{ij}^k$, are then Fourier-encoded~\cite{yifan2021input} to produce geometric embeddings $\mathcal{E}_{o}$, $\mathcal{E}_r$ and $\mathcal{E}_x$, with a mapping of:
\begin{equation}
\small
x \mapsto 
\left[
x, \sin(f_1\pi x), \cos(f_1\pi x), \dots, \sin(f_M\pi x), \cos(f_M\pi x) 
\right]
\end{equation}
where $M$ is the number of Fourier frequencies used ($M_o$ for camera origin, $M_r$ for viewing ray, and $M_x$ for 3D point), equally spaced in the interval $[1,\frac{\mu}{2}]$. 
Here, $\mu$ is a maximum frequency parameter shared across all dimensions. 
These embeddings are concatenated to be used as queries by the cross-attention decoders described below. 
Ray embeddings are defined as $\mathcal{E}_{ray} = \mathcal{E}_{o} \oplus \mathcal{E}_{r}$ and volumetric embeddings as $\mathcal{E}_{vol} = \mathcal{E}_{o} \oplus \mathcal{E}_{x}$, where $\oplus$ denotes concatenation.

\subsection{Cross-Attention Decoders}
\label{sec:decoders}

We use task-specific decoders, with one cross-attention layer between the $N_q \times C_q$ queries and the $N_l \times C_l$ latent space $\mathcal{S}$ followed by an MLP that produces a $N_q \times C_o$ output (more details in the supplementary material). 
This output is processed to generate estimates as described below. 

\noindent\textbf{The radiance head} 
$\mathcal{H}_R$ decodes volumetric embeddings $\mathcal{E}_{vol}$ as a $4$-dimensional vector $(\textbf{c},\sigma)$, where $\textbf{c}=(r,g,b)$ are colors and $\sigma$ is density. 
A sigmoid is used to normalize colors between $[0,1]$, and a ReLU truncates densities to positive values. 
To generate per-pixel predictions, we composite $K$ predictions along its viewing ray~\cite{mildenhall2020nerf}, using sampled depth values $Z_{ij} = \{z_k\}_{k=0}^{K-1}$. 
The resulting per-pixel predicted color $\hat{\textbf{c}}_{ij}$ and depth $\hat{d}_{ij}$ is given by:
\begin{equation}
\small
\hat{\textbf{c}}_{ij} = \sum_{k=1}^K w_k\hat{\textbf{c}}_k
\quad , \quad
\hat{d}_{ij} = \sum_{k=1}^{K} w_k z_k 
\label{eq:vol_pred}
\end{equation}
Per-point weights $w_k$ and accumulated densities $T_k$, given intervals $\delta_k = z_{k+1} - z_k$, are defined as:
\begin{align}
\small
w_k &= T_k \Big( 1-\exp(-\sigma_k\delta_k)
\Big) 
\vspace{-1mm}
\\
T_k &= \exp \Big(
-\sum_{k'=1}^K \sigma_{k'}\delta_{k'} 
\Big) 
\end{align}
\noindent\textbf{The light field head} 
$\mathcal{H}_L$ decodes ray embeddings $\mathcal{E}_{ray}$ as a $3$-dimensional vector $\hat{\textbf{c}}_{ij}=(r,g,b)$ containing pixel colors. 
These values are normalized between $[0,1]$ with a sigmoid.

\noindent\textbf{The depth field head}
$\mathcal{H}_D$  decodes ray embeddings $\mathcal{E}_{ray}$ as a scalar value $\hat{d}_{ij}$ representing predicted pixel depth.
This value is normalized between a $[d_{min},d_{max}]$ range.

\subsection{Self-Supervised Losses}
\label{sec:self-supervised}

We combine the traditional volumetric rendering view synthesis loss $\mathcal{L}_s$ (\Section \ref{sec:view_loss}) with the multi-view photometric loss $\mathcal{L}_p$ (\Section \ref{sec:photometric_loss}), using $\alpha_p$ as a weight coefficient:
\begin{equation}
\mathcal{L} = \mathcal{L}_s + \alpha_p\mathcal{L}_p
\end{equation}
\subsubsection{Single-View Volumetric Rendering}
\label{sec:view_loss}

We use Mean Squared Error (MSE) to supervise the predicted image $\hat{I}_t$ (where $\hat{I}_{t}(i,j) = \hat{\textbf{c}}_{ij}$, see \Equation \ref{eq:vol_pred}), relative to the target image $I_t$. 
\begin{equation}
\mathcal{L}_{s} = ||\hat{I}_t - I_t||^2    
\label{eq:view_loss}
\end{equation}
This is the standard objective for radiance-based reconstruction~\cite{mildenhall2020nerf}. Importantly, this is a \textit{single-view} objective, since it directly compares prediction and ground truth without considering additional viewpoints. 
Therefore, multi-view consistency must be learned implicitly by observing the same scene from multiple viewpoints.
When such information is not available (e.g., forward-facing datasets with limited viewpoint diversity), it may lead to degenerate geometries that do not properly model the observed 3D space.

\begin{figure*}[t!] 
\vspace{-2mm}
  \setlength{\tabcolsep}{1.5mm}
  \begin{tabular}{cccc}
  \small{Ground Truth} & \small{Baseline (radiance field)} & \small{DeLiRa (radiance field)} & \small{DeLiRa (depth and light fields)} \\
  \cmidrule(lr){1-1} \cmidrule(lr){2-2} \cmidrule(lr){3-3} \cmidrule(lr){4-4}
  \includegraphics[width=0.13\textwidth,height=1.4cm]{
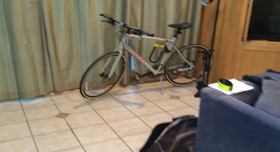} &
  \includegraphics[width=0.13\textwidth,height=1.4cm]{
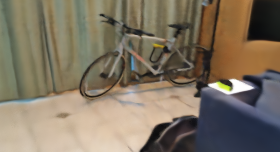}
  \includegraphics[width=0.13\textwidth,height=1.4cm]{
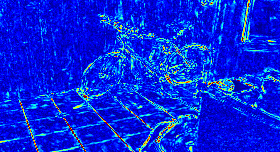} &
  \includegraphics[width=0.13\textwidth,height=1.4cm]{
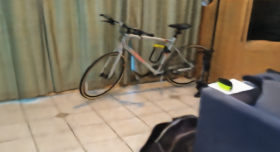}
  \includegraphics[width=0.13\textwidth,height=1.4cm]{
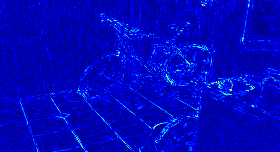} &
  \includegraphics[width=0.13\textwidth,height=1.4cm]{
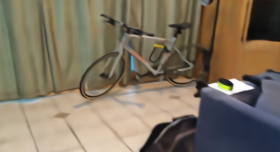}
  \includegraphics[width=0.13\textwidth,height=1.4cm]{
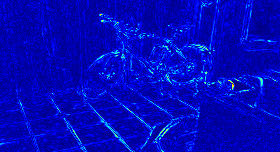} 
\vspace{-1mm}
\\ 
\includegraphics[width=0.13\textwidth,height=1.4cm]{
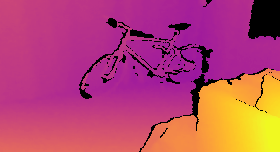} &
\includegraphics[width=0.13\textwidth,height=1.4cm]{
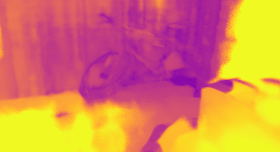}
\includegraphics[width=0.13\textwidth,height=1.4cm]{
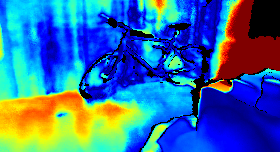} &
\includegraphics[width=0.13\textwidth,height=1.4cm]{
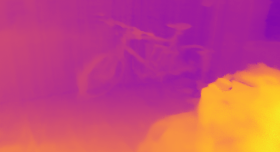}
\includegraphics[width=0.13\textwidth,height=1.4cm]{
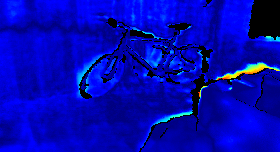} &
\includegraphics[width=0.13\textwidth,height=1.4cm]{
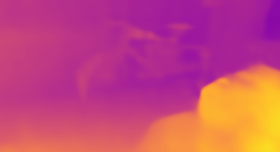}
\includegraphics[width=0.13\textwidth,height=1.4cm]{
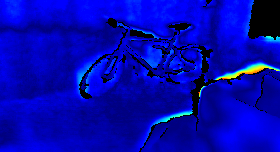} 
\end{tabular}
\vspace{-4mm}
\caption{
\textbf{Qualitative depth and view synthesis results} from unseen viewpoints, using different \Acronym decoders. As a baseline, we show predictions obtained from a model trained without our contributions, leading to a degenerate learned geometry due to shape-radiance ambiguity (i.e., accurate view synthesis with poor depth predictions).
\label{fig:qualitative}
}
\vspace{-5mm}
\end{figure*}

\subsubsection{Multi-View Photometric Warping}
\label{sec:photometric_loss}

To address this limitation, we introduce the self-supervised multi-view photometric objective~\cite{monodepth2, packnet} as an additional source of self-supervision in the volumetric rendering setting. 
For each pixel $(u,v)$ in target image $I_t$, with predicted depth $\hat{d}$ (e.g., see \Equation \ref{eq:vol_pred}), we obtain the projected coordinates $(u',v')$ with predicted depth $\hat{d'}$ in a context image $I_c$ via a warping operation, defined as:
\begin{equation}
 \small
    \hat{d'} \left[\begin{matrix}u' \\ v' \\ 1 \\
    \end{matrix}\right] = 
    \mathbf{K}_c \mathbf{R}_{t \rightarrow c} \left( \mathbf{K}^{-1}_t 
    \left[\begin{matrix}u \\ v \\ 1 \\ 
    \end{matrix}\right] \hat{d} 
    + \mathbf{t}_{t \rightarrow c}\right)
\end{equation}

To produce a synthesized target image, we use grid sampling with bilinear interpolation~\cite{jaderberg2015spatial} to place information from the context image onto each target pixel, given their corresponding warped coordinates.
The photometric reprojection loss between target $I_t$ and synthesized $\hat{I}_t$ images consists of a weighted sum with a structure similarity (SSIM) term~\cite{wang2004image} and an L1 loss term:
\begin{equation}
\small
\mathcal{L}_{p}(I_t,\hat{I_t}) = \alpha~\frac{1 - \text{SSIM}(I_t,\hat{I_t})}{2} + (1-\alpha)~\| I_t - \hat{I_t} \|
\label{eq:photo_loss}
\end{equation}

\noindent\textbf{Strided ray sampling} 
Due to the large amount of network calls required for volumetric rendering, it is customary to randomly sample rays at training time~\cite{mildenhall2020nerf}. 
This is possible because the volumetric view synthesis loss (\Equation \ref{eq:view_loss}) can be calculated at a per-pixel basis. 
The photometric loss (\Equation \ref{eq:photo_loss}), however, requires a dense image, and thus is incompatible with random sampling. 
To circumvent this, while maintaining reasonable training times and memory usage, we use \textit{strided ray sampling}.
Fixed horizontal $s_w$ and vertical $s_h$ strides are used, and random horizontal $o_w \in [0,s_w-1]$ and vertical $o_h \in [0,s_h-1]$ offsets are selected to determine the starting point of the sampling process, resulting in $s_h \times s_w$ combinations. 
The rays can be arranged to produce a downsampled image $I'_t$ of resolution $\lfloor\frac{H - o_h}{s_h}\rfloor \times \lfloor\frac{W - o_w}{s_w}\rfloor$, with corresponding predicted image $\hat{I}'_t$ and depth map $\hat{D}'_t$. 
Note that the target intrinsics $\textbf{K}'$ have to be adjusted accordingly, and context images do not need to be downsampled.

\subsection{Light and Depth Field Decoding}

We take advantage of the general nature of our framework to produce novel view synthesis and depth estimates in two different ways: \textit{indirectly}, by compositing predictions from a volumetric decoder (\Equation \ref{eq:vol_pred}); and \textit{directly}, as the output of light and depth field decoders. 
Because light and depth field predictions~\cite{tri-define} lack the multi-view consistency inherent to volumetric rendering~\cite{tri-define,wang2022r2l,sitzmann2021light}, we augment the amount of available training data by including virtual supervision from volumetric predictions.

This is achieved by randomly sampling virtual cameras from novel viewpoints at training time, and using volumetric predictions as pseudo-labels to supervise the light and depth field predictions. 
Virtual cameras are generated by adding translation noise $\bm{\epsilon}_v = [\epsilon_x,\epsilon_y,\epsilon_z]_v \sim \mathcal{N}(0,\sigma_v)$ to the pose of an available camera, selected randomly from the training dataset. 
The viewing angle is set to point towards the center of the original camera, at a distance of $d_{max}$, which is also disturbed by $\bm{\epsilon}_c = [\epsilon_x,\epsilon_y,\epsilon_z]_c \sim \mathcal{N}(0,\sigma_v)$. 
We can use information from this virtual camera to decode a predicted volumetric image $\hat{I}_v$ and depth map $\hat{D}_v$, as well as a predicted image  $\hat{I}_l$ and depth map $\hat{D}_d$ from the light and depth field decoders. 
We use the MSE loss (\Section \ref{sec:view_loss}) to supervise $\hat{I}_l$ relative to $\hat{I}_v$, as well as the L1-log loss to supervise $\hat{D}_d$ relative to $\hat{D}_v$, resulting in the virtual loss:
\begin{equation}
\small
\mathcal{L}_v = 
\left\Vert 
\log(\hat{D}_{r}) - \log(\hat{D}_{d})
\right\Vert_1 + 
\left(
\hat{I}_{r} - \hat{I}_l 
\right)^2
\end{equation} 
Note that the self-supervised losses from \Section \ref{sec:self-supervised} are also applied to the original light and depth field predictions.

\subsection{Depth Field Volumetric Guidance}
\label{sec:guidance}

In the previous section we described how volumetric predictions can be used to improve light and depth field estimates, by introducing additional supervision in the form of virtual cameras. 
Here, we show how depth field predictions can be used to improve the efficiency of volumetric estimates, by sampling from areas near the observed surface. 
Although more involved strategies have been proposed~\cite{dsnerf,depthpriorsnerf,donerf}, we found that sampling from a Gaussian distribution $\mathcal{N}(\hat{D}_d,\sigma_g)$, centered around $\hat{D}_d$ with standard deviation $\sigma_g$, provided optimal results.
Importantly, all these strategies require additional information from pre-trained depth networks or sparse supervision, whereas ours use predictions generated by the same network, learned from scratch and decoded from the same representation. 

\subsection{Training Schedule}
\label{sec:schedule}

Since all predictions are learned jointly, we use a training schedule such that depth field estimates can reach a reasonable level of performance before serving as guidance for volumetric sampling. 
In the first 400 epochs ($10\%$ of the total number of steps), we consider $K$ ray samples, and depth field guidance (DFG) is not used. Afterwards, $K_g$ samples are relocated to DFG, and drawn instead from $\mathcal{N}(\hat{D}_d,\sigma_g)$. After another 400 epochs, we once again reduce the amount of ray samples by $K_g$, but this time without increasing the number of depth field samples, which decreases the total number of samples used for volumetric rendering. This process is repeated every 400 epochs, and at $K=0$ ray sampling is no longer performed, only DFG with $K_g$ samples. 

Moreover, we note that the multi-view photometric objective is unable to model view-dependent artifacts, since it relies on explicit image warping. Thus, we gradually remove this regularization, so that our network can first converge to the proper implicit scene geometry, and then use it to further improve novel view synthesis. In practice, after every $400$ epochs we decay $\mathcal{L}_p$ by a factor of $0.8$, and completely remove it in the final $800$ epochs.

\section{Experimental Results}

\captionsetup[table]{skip=6pt}

\begin{table*}[t!]
\small
\renewcommand{\arraystretch}{0.85}
\centering
{
\small
\setlength{\tabcolsep}{0.4em}
\begin{tabular}{lc|c||c|c|c|c||c|c|c}
\toprule
\multicolumn{2}{l|}{\multirow{2}[2]{*}{\textbf{Method}}} &
\multirow{2}[2]{*}{\rotatebox{90}{\scriptsize{Superv.}}} &
\multicolumn{4}{c||}{\textit{Lower is better}} &
\multicolumn{3}{c}{\textit{Higher is better}} 
\\
\cmidrule(lr){4-7} \cmidrule(lr){8-10}
 & & & Abs.Rel. &
Sq.Rel. &
RMSE &
RMSE$_{log}$ &
$\delta < {1.25}$ &
$\delta < {1.25}^2$ &
$\delta < {1.25}^3$
\\
\toprule
\multicolumn{2}{l|}{COLMAP~\cite{colmap}} & --
& $0.462$ & $0.631$ & $1.012$ & $1.734$ & $0.481$ & $0.514$ & $0.533$ \\
\midrule
\multicolumn{2}{l|}{ACMP~\cite{Xu2020ACMP}} & D
& $0.194$ & $0.171$ & $0.455$ & $0.306$ & $0.731$ & $0.881$ & $0.942$ \\
\multicolumn{2}{l|}{DELTAS~\cite{sinha2020deltas}} & D
& $0.100$ & $0.032$ & $0.207$ & $0.128$ & $0.862$ & $0.992$ & $0.999$ \\
\multicolumn{2}{l|}{DeepV2D~\cite{deepv2d}} & D 
& $0.082$ & $0.023$ & $0.171$ & $0.109$ & $0.941$ & $0.991$ & $0.998$ \\
\multicolumn{2}{l|}{Atlas~\cite{murez2020atlas}} & D
& $0.078$ & $0.063$ & $0.244$ & $0.269$ & $0.929$ & $0.954$ & $0.959$ \\
\midrule
\multicolumn{2}{l|}{NeRF~\cite{mildenhall2020nerf}} & --
& $0.393$ & $1.485$ & $1.090$ & $0.521$ & $0.489$ & $0.732$ & $0.828$ \\
\multicolumn{2}{l|}{CVD~\cite{luo2020consistent}} & -- 
& $0.099$ & $0.030$ & $0.194$ & $0.127$ & $0.901$ & $0.988$ & $0.997$ \\
\multicolumn{2}{l|}{NerfingMVS~\cite{nerfingmvs} (w/o filter)} & C
& $0.063$ & \best $\textbf{0.014}$ & $0.145$ & $0.094$ & $0.954$ & $0.991$ & \sbest $\underline{0.999}$ \\
\multicolumn{2}{l|}{NerfingMVS~\cite{nerfingmvs} (w/ filter)} & C
& $0.061$ & \best $\textbf{0.014}$ & \sbest $\underline{0.134}$ & \sbest $\underline{0.086}$ & $0.960$ & \best $\textbf{0.995}$ & \sbest $\underline{0.999}$ \\
\midrule
\midrule
\multirow{0}[2]{*}{\textbf{\Acronym}} & R & -- 
& \sbest $\underline{0.055}$ & \best $\textbf{0.014}$ & \best $\textbf{0.131}$ & \best $\textbf{0.082}$ & \best $\textbf{0.970}$ & \sbest $\underline{0.994}$ & \sbest $\underline{0.999}$ \\
& D & -- 
& \best $\textbf{0.054}$ & \sbest $\underline{0.028}$ & $0.138$ & $0.088$ & \sbest $\underline{0.966}$ & $0.992$ & \best $\textbf{1.000}$ \\
\bottomrule
\end{tabular}
}
\caption{
\textbf{Average depth synthesis results} on \emph{ScanNet-Frontal}. \emph{Superv.} indicates the source of depth supervision: \emph{D} for ground truth, and \emph{C} for COLMAP predictions. \Acronym outperforms all other methods, despite not requiring supervision. Moreover, our depth field results (D) are on par with radiance predictions (R), and can be generated with a single query.
}
\label{table:depth_frontal}
\vspace{-1mm}
\end{table*}

\captionsetup[table]{skip=6pt}

\begin{table*}[t!]
\small
\renewcommand{\arraystretch}{0.9}
\centering
{
\small
\setlength{\tabcolsep}{0.25em}
\begin{tabular}{lc|cc|cc|cc|cc|cc|cc|cc|cc}
\toprule
\multicolumn{2}{l|}{\multirow{2}[2]{*}{\textbf{Method}}} & 
\multicolumn{2}{c|}{\textit{Scene 0000}} &
\multicolumn{2}{c|}{\textit{Scene 0079}} &
\multicolumn{2}{c|}{\textit{Scene 0158}} &
\multicolumn{2}{c|}{\textit{Scene 0316}} &
\multicolumn{2}{c|}{\textit{Scene 0521}} &
\multicolumn{2}{c|}{\textit{Scene 0553}} &
\multicolumn{2}{c|}{\textit{Scene 0616}} &
\multicolumn{2}{c}{\textit{Scene 0653}}
\\
& & PSNR & SSIM & PSNR & SSIM & PSNR & SSIM & PSNR & SSIM
& PSNR & SSIM & PSNR & SSIM & PSNR & SSIM & PSNR & SSIM \\
\toprule
\multicolumn{2}{l|}{NSVF~\cite{liu2020neural}}
& $23.36$ & $0.823$ & $26.88$ & $0.887$ & $31.98$ & $0.951$ & $22.29$ & $0.917$ & $27.73$ & $0.892$ & $31.15$ & $0.947$ & $15.71$ & $0.704$ & $28.95$ & $0.929$ 
\\
\multicolumn{2}{l|}{SVS~\cite{Riegler2021SVS}}        
& $21.39$ & \sbest $\underline{0.914}$ & $25.18$ & \best $\textbf{0.923}$ & $29.43$ & $0.953$ & $20.63$ & \sbest $\underline{0.941}$ & $27.97$ & \best $\textbf{0.924}$ & $30.95$ & $0.968$ & \best $\textbf{21.38}$ & \best $\textbf{0.899}$ & $27.91$ & $0.965$
\\
\multicolumn{2}{l|}{NeRF~\cite{mildenhall2020nerf}}
& $18.75$ & $0.751$ & $25.48$ & $0.896$ & $29.19$ & $0.928$ & $17.09$ & $0.828$ & $24.41$ & $0.871$ & $30.76$ & $0.950$ & $15.76$ & $0.699$ & $30.89$ & $0.953$ 
\\
\multicolumn{2}{l|}{NerfingMVS~\cite{nerfingmvs}}
& $22.10$ & $0.880$ & $27.27$ & $0.916$ & $30.55$ & $0.948$ & $20.88$ & $0.899$ & $28.07$ & $0.901$ & $32.56$ & $0.965$ & $18.07$ & $0.748$ & $31.43$ & $0.964$ 
\\
\midrule
\multirow{0}[2]{*}{\textbf{\Acronym}}
& R & \best $\textbf{25.88}$ & \best $\textbf{0.919}$
& \sbest $\underline{28.01}$ & $0.916$ 
& \sbest $\underline{34.68}$ & \sbest $\underline{0.969}$ 
& \best $\textbf{23.31}$ & \best $\textbf{0.948}$ 
& \sbest $\underline{28.97}$ & $0.909$ 
& \best $\textbf{36.32}$ & \best $\textbf{0.981}$ 
& \sbest $\underline{20.27}$ & \sbest $\underline{0.851}$ 
& \sbest $\underline{33.70}$ & \sbest $\underline{0.967}$ 
\\
& L & \sbest $\underline{25.34}$ & $0.907$ 
& \best$\textbf{28.48}$ & \sbest $\underline{0.920}$
& \best$\textbf{35.77}$ & \best$\textbf{0.980}$ 
& \sbest$\underline{23.18}$ & $0.937$
& \best$\textbf{29.22}$ & \sbest $\underline{0.916}$
& \sbest$\underline{35.94}$ & \sbest $\underline{0.974}$ 
& $19.18$ & $0.832$ 
& \best$\textbf{34.63}$ & \best$\textbf{0.973}$ 
\\
\bottomrule
\end{tabular}
}
\caption{
\textbf{Per-scene view synthesis results}, on \emph{ScanNet-Frontal}. \Acronym improves view synthesis results (PSNR) in all considered scenes ($+9.8\% \pm 4.1\%$), relative to the previous state of the art~\cite{nerfingmvs}. 
}
\label{table:view_frontal}
\vspace{-3mm}
\end{table*}

\captionsetup[table]{skip=6pt}

\begin{table}[t!]
\small
\renewcommand{\arraystretch}{0.85}
\centering
{
\small
\setlength{\tabcolsep}{0.3em}
\begin{tabular}{lc|c||c||c|c|c}
\toprule
\multicolumn{2}{l|}{\multirow{2}[2]{*}{\textbf{Method}}} & 
\multirow{2}[2]{*}{\rotatebox{90}{\scriptsize{Superv.}}} &
\multicolumn{1}{c||}{\textit{Depth}} &
\multicolumn{3}{c}{\textit{View Synthesis}}
\\
\cmidrule(lr){4-4} \cmidrule(lr){5-7} 
& & &
RMSE$\downarrow$ &
PSNR$\uparrow$ & 
SSIM$\uparrow$ &
LPIPS$\downarrow$
\\
\toprule
\multicolumn{2}{l|}{NeRF~\cite{mildenhall2020nerf}}
& - & $1.163$ & $19.03$ & $0.670$ & $0.398$ \\
\multicolumn{2}{l|}{DS-NeRF~\cite{dsnerf}} 
& C & $0.423$ & $20.94$ & $0.721$ & $0.330$ \\
\multicolumn{2}{l|}{NerfingMVS~\cite{nerfingmvs}$^\dagger$} 
& C & $0.469$ & $16.45$ & $0.641$ & $0.488$ \\
\multicolumn{2}{l|}{DDP-NeRF~\cite{depthpriorsnerf}$^\dagger$} 
& C & $0.504$ & $20.71$ & $0.719$ & $0.337$ \\
\multicolumn{2}{l|}{DDP-NeRF~\cite{depthpriorsnerf}$^\dagger$} 
& D+C & $0.229$ & $21.02$ & $0.742$ & \best $\textbf{0.289}$ \\
\midrule
\multirow{0}[2]{*}{\textbf{\Acronym}} & R & - & \sbest $\underline{0.215}$ & \best $\textbf{21.64}$ & \best $\textbf{0.761}$ & \sbest $\underline{0.302}$ \\
& DL & - & \best $\textbf{0.213}$ & \sbest $\underline{21.26}$ & \sbest $\underline{0.748}$ & $0.305$ 
\\
\bottomrule
\end{tabular}
}
\caption{
\textbf{Depth and view synthesis results} on \emph{ScanNet-Rooms}. The \emph{superv.} column indicates the source of depth supervision: D denotes ground-truth depth maps, and C denotes COLMAP predictions. The symbol $^\dagger$ indicates the use of separate depth networks, pre-trained on additional data. 
}
\label{table:depth_rooms}
\vspace{-5mm}
\end{table}

\subsection{Dataset}

The primary goal of \Acronym is to enable novel view synthesis in the limited viewpoint diversity setting, which is very challenging for implicit representations (prior work on few-shot NeRF~\cite{dietnerf, niemeyer2022regnerf} is limited to synthetic or tabletop settings). Thus, following standard protocol~\cite{dsnerf,nerfingmvs,depthpriorsnerf} we use the ScanNet dataset~\cite{dai2017scannet} as our evaluation benchmark. This is a challenging benchmark, composed of real-world room-scale scenes subject to motion blur and noisy calibration~\cite{depthpriorsnerf}. 
For a fair comparison with other methods, we consider two different training and testing splits: \textit{ScanNet-Frontal}~\cite{nerfingmvs}, composed of $8$ scenes, and \textit{ScanNet-Rooms}~\cite{depthpriorsnerf}, composed of $3$ scenes. 
For more details, please refer to the supplementary material. 

\subsection{Volumetric Depth and View Synthesis}

First, we evaluate the performance of \Acronym focusing on volumetric predictions.  Improvements in depth synthesis are expected to lead to improvements in view synthesis, which we validate in the following section. 
In \Table \ref{table:depth_frontal} we compare our depth synthesis results on \textit{ScanNet-Frontal} with several classical methods~\cite{Xu2020ACMP,sinha2020deltas,deepv2d,murez2020atlas}, all of which require ground-truth depth maps as a source of supervision. 
We also consider NeRF~\cite{mildenhall2020nerf}, that only optimizes for volumetric rendering, as well as  CVD~\cite{luo2020consistent} and NerfingMVS\cite{nerfingmvs}, that use a pre-trained depth network fine-tuned on in-domain sparse information.
Even though \Acronym operates in the same setting as NeRF (i.e., only posed images), it still achieves substantial improvements over all other methods. 
In particular, \Acronym improves upon NeRF by $86.3\%$ in absolute relative error (AbsRel) and $88.0\%$ in root mean square error (RMSE), as well as improving upon the previous state of the art~\cite{nerfingmvs} by $14.2\%$ in Abs.Rel. and $9.6\%$ in RMSE. 
Similar trends are also observed for view synthesis, where \Acronym improves upon \cite{nerfingmvs} in PSNR by $11.1\%$, as shown in \Table \ref{table:view_frontal}. 
We attribute this to the fact that our regularization leverages dense direct self-supervision from the environment, rather than relying on sparse noisy samples to fine-tune a pre-trained network.

\captionsetup[table]{skip=6pt}

\begin{table}[t!]
\small
\renewcommand{\arraystretch}{0.85}
\centering
\small
\setlength{\tabcolsep}{0.3em}
\begin{tabular}{c|l|c||c|c||c|c}
\toprule
& \multirow{2}[2]{*}{\textbf{Version}} &
\multirow{2}[2]{*}{\rotatebox{90}{\scriptsize{Decoder}}} & 
\multicolumn{2}{c||}{\textbf{Depth}$\downarrow$}  & 
\multicolumn{2}{c}{\textbf{View Synth.}$\uparrow$}
\\
\cmidrule(lr){4-5} \cmidrule(lr){6-7} 
\renewcommand{\arraystretch}{1.0}
& & &
AbsRel &
RMSE &
PSNR & 
SSIM
\\
\toprule
A & \footnotesize{Monodepth (self-sup.)} & - 
& $0.096$ & $0.268$ & --- & --- 
\\
B & \footnotesize{COLMAP sup.} & \scriptsize{R}  
& $0.134$ & $0.401$ & $26.17$ & $0.862$ 
\\
\midrule
C & \footnotesize{DeLiRa (w/o self-sup.)} & \scriptsize{R}
& $0.245$ & $0.570$ & $27.64$ & $0.895$ 
\\
D & \footnotesize{MLP (w/o self-sup.)} & \scriptsize{R}
& $0.238$ & $0.546$ & $27.50$ & $0.882$ 
\\
E & \footnotesize{MLP (self-sup.)} & \scriptsize{R}
& $0.068$ & $0.163$ & $27.51$ & $0.887$ 
\\
F & \footnotesize{DeLiRa (1-MLP)} & \scriptsize{DL}
& $0.057$ & $0.133$ & $27.34$ & $0.889$ 
\\
\midrule
G & \footnotesize{Volumetric-only} & \scriptsize{R}
& $0.059$ & $0.142$ & $27.87$ & $0.891$ 
\\
H & \footnotesize{Depth / Light-only} & \scriptsize{DL}
& $0.123$ & $0.344$ & $23.56$ & $0.757$ 
\\
I & \footnotesize{Distilled (separate $\mathcal{S}_{DL}$)} & \scriptsize{DL}
& $0.063$ & $0.152$ & $27.52$ & $0.881$ 
\\
J & \footnotesize{Distilled (shared $\mathcal{S}_{R}$)} & \scriptsize{DL}
& $0.066$ & $0.169$ & $27.33$ & $0.868$ 
\\
K & \footnotesize{Distilled (shared $\mathcal{S}_{R}$,\st{VCA})} & \scriptsize{DL}
& $0.076$ & $0.181$ & $26.16$ & $0.849$ 
\\
\midrule
& & \scriptsize{R}
& $0.058$ & $0.140$ & $28.04$ & $0.898$ 
\\
\multirow{-2}{*}{L}
& \multirow{-2}{*}{\footnotesize{Separate $\mathcal{S}_R$ and $\mathcal{S}_{DL}$}} & \scriptsize{DL}
& $0.061$ & $0.152$ & $27.77$ & $0.892$ 
\\
& & \scriptsize{R}
& $0.077$ & $0.210$ & $27.74$ & $0.877$ 
\\
\multirow{-2}{*}{M} 
& \multirow{-2}{*}{\footnotesize{Without VCA}} & \scriptsize{DL}
& $0.079$ & $0.227$ & $27.25$ & $0.855$ 
\\
& & \scriptsize{R}
& $0.059$ & $0.145$ & $28.27$ & $0.908$ 
\\
\multirow{-2}{*}{N}
& \multirow{-2}{*}{\footnotesize{Without DFG}} & \scriptsize{DL}
& $0.055$ & $0.139$ & $28.73$ & $0.924$ 
\\
& & \scriptsize{R}
& $0.056$ & $0.135$ & $28.59$ & $0.922$ 
\\
\multirow{-2}{*}{O}
& \multirow{-2}{*}{\footnotesize{Without vanishing $\mathcal{L}_p$}} & \scriptsize{DL}
& $0.055$ & $0.144$ & $28.74$ & $0.918$ 
\\
\midrule
& & \scriptsize{R} 
& $0.055$ & \best $\mathbf{0.131}$ & $28.98$ & $0.934$ 
\\
\multirow{-2}{*}{} 
& \multirow{-2}{*}{\textbf{\Acronym}} & \scriptsize{DL}
& \best $\mathbf{0.054}$ & $0.138$ & \best $\mathbf{29.10}$ & \best $\mathbf{0.932}$
\\
\bottomrule
\end{tabular}
\caption{
\textbf{Ablative analysis} of the various components of our proposed \Acronym method, on \textit{ScanNet-Frontal}. \emph{R}, \emph{D}, and \emph{L} indicate radiance, depth, and light field predictions. 
}
\label{table:nmvs_ablation}
\vspace{-4mm}
\end{table}

In \Table \ref{table:depth_rooms} we show a similar evaluation on \textit{ScanNet-Rooms}, which constitutes a more challenging setting due to a larger area coverage (an entire room, as opposed to a local region). In fact, as shown in \cite{depthpriorsnerf}, most methods struggle in this setting: NeRF~\cite{mildenhall2020nerf} generates ``floaters" due to limited viewpoint diversity, DS-NeRF~\cite{dsnerf} is prone to errors in sparse depth input, and the error map calculation of NerfingMVS~\cite{nerfingmvs} fails when applied to larger areas. DDP-NeRF~\cite{depthpriorsnerf} circumvents these issues by using an additional uncertainty-aware depth completion network, trained on RGB-D data from the same domain. Even so, our simpler approach of regularizing the volumetric depth using a multi-view photometric objective leads to a new state of the art, both in depth and novel view synthesis. 

\subsection{Light and Depth Field Performance}

In addition to volumetric rendering, \Acronym also generates light and depth field predictions, that can be efficiently decoded with a single network forward pass. 
We report these results in the same benchmarks, achieving state-of-the-art performance comparable to their corresponding volumetric predictions. 
In the supplementary material we explore the impact of using different decoder architectures, noticing that deeper networks yield significant improvements in view synthesis, which is in agreement with \cite{wang2022r2l}.
Interestingly, we did not observe similar improvements in depth synthesis, which we attribute to the lack of higher frequency details in this task.

\subsection{Ablative Analysis}

\begin{figure}[t!]
  \setlength{\tabcolsep}{1.5mm}
  \begin{tabular}{cc}
  \small{Radiance field} & \small{Depth and light fields} \\
  \cmidrule(lr){1-1} \cmidrule(lr){2-2}
\includegraphics[width=0.2\textwidth
,trim={10.0cm 4.5cm 10.0cm 3.0cm},clip
]{
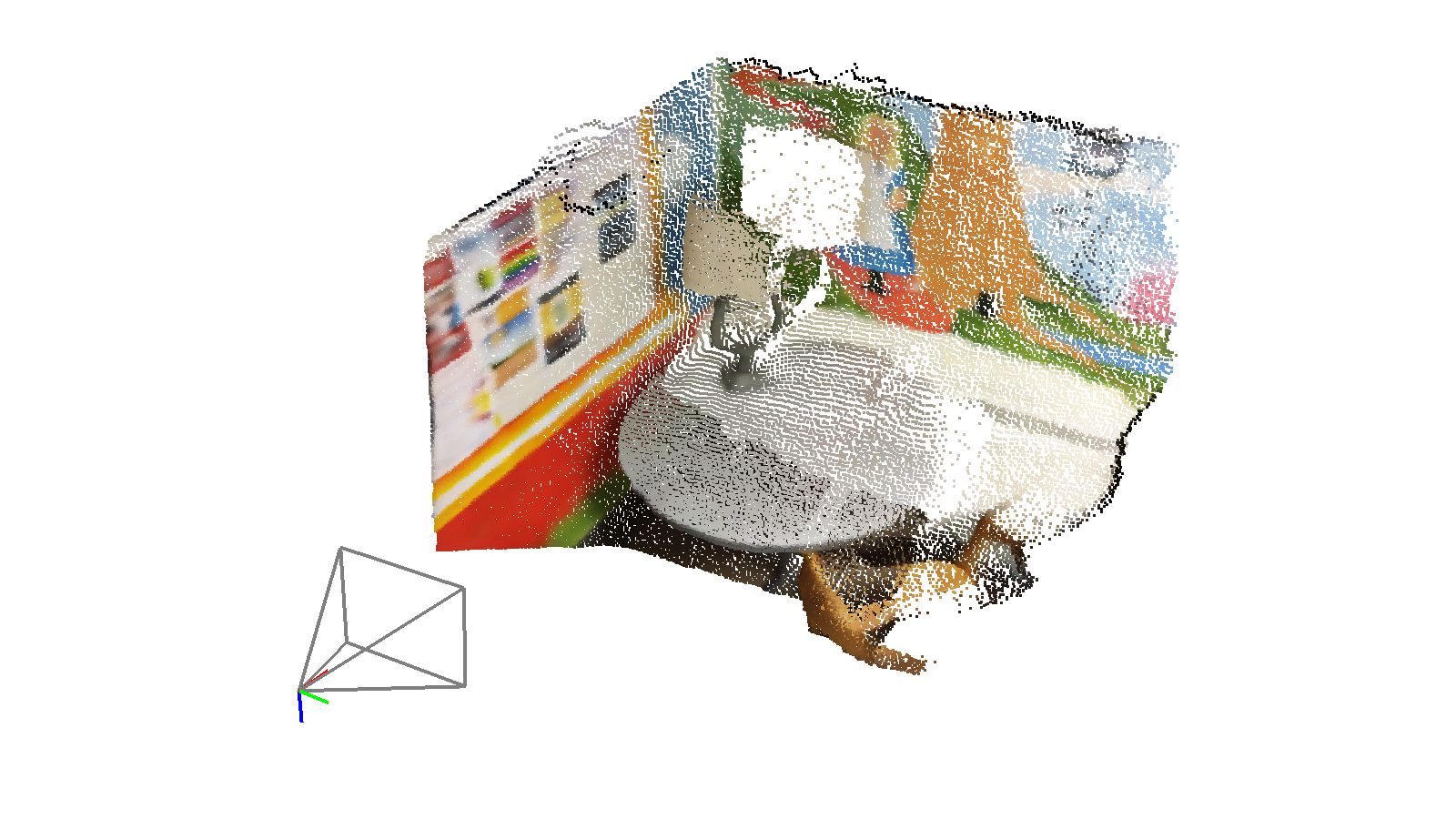}
&
\includegraphics[width=0.2\textwidth
,trim={10.0cm 4.5cm 10.0cm 3.0cm},clip
]{
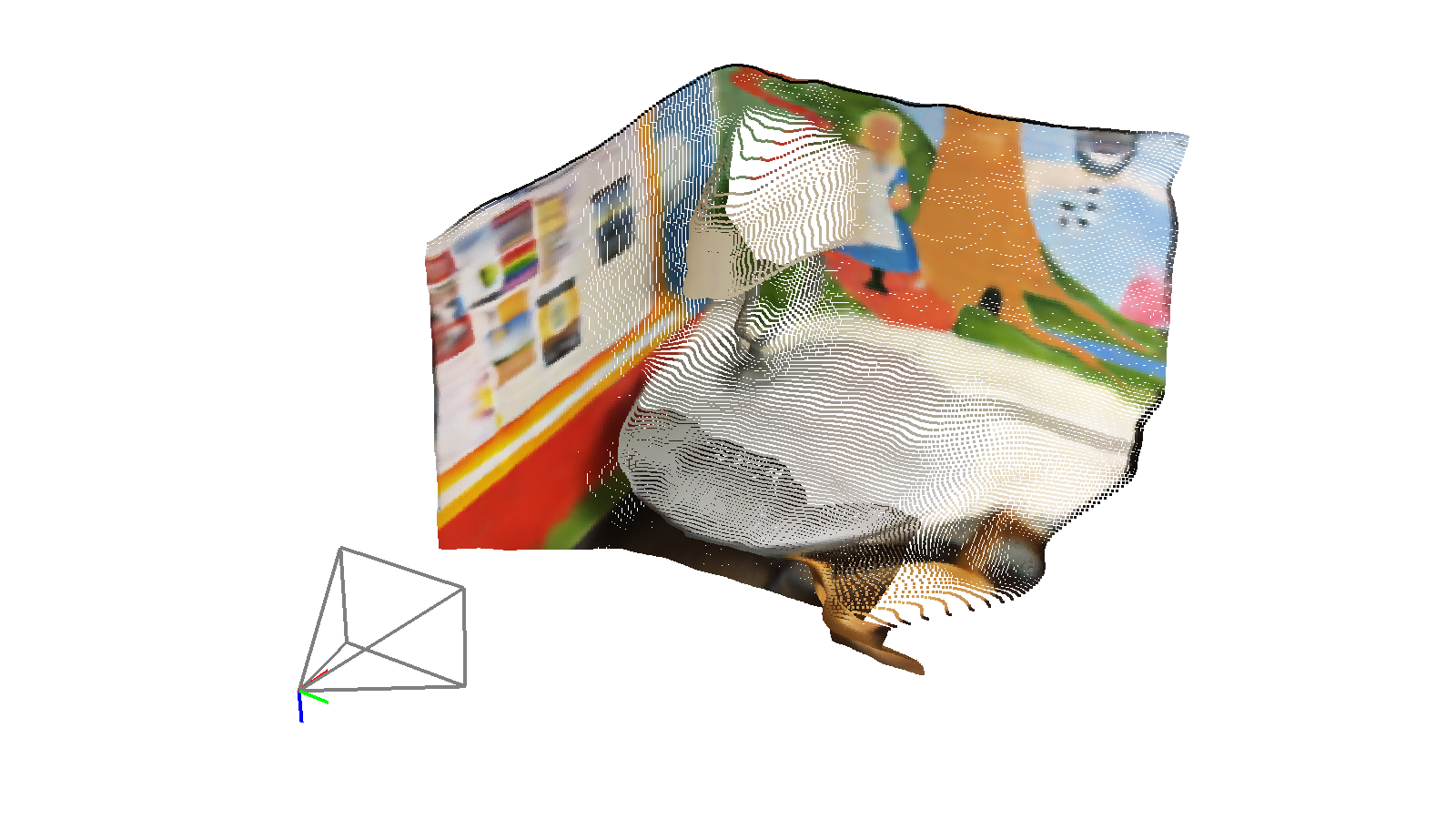}
\vspace{-1mm}
\\
\includegraphics[width=0.2\textwidth
,trim={5.0cm 5.0cm 10.0cm 1.0cm},clip
]{
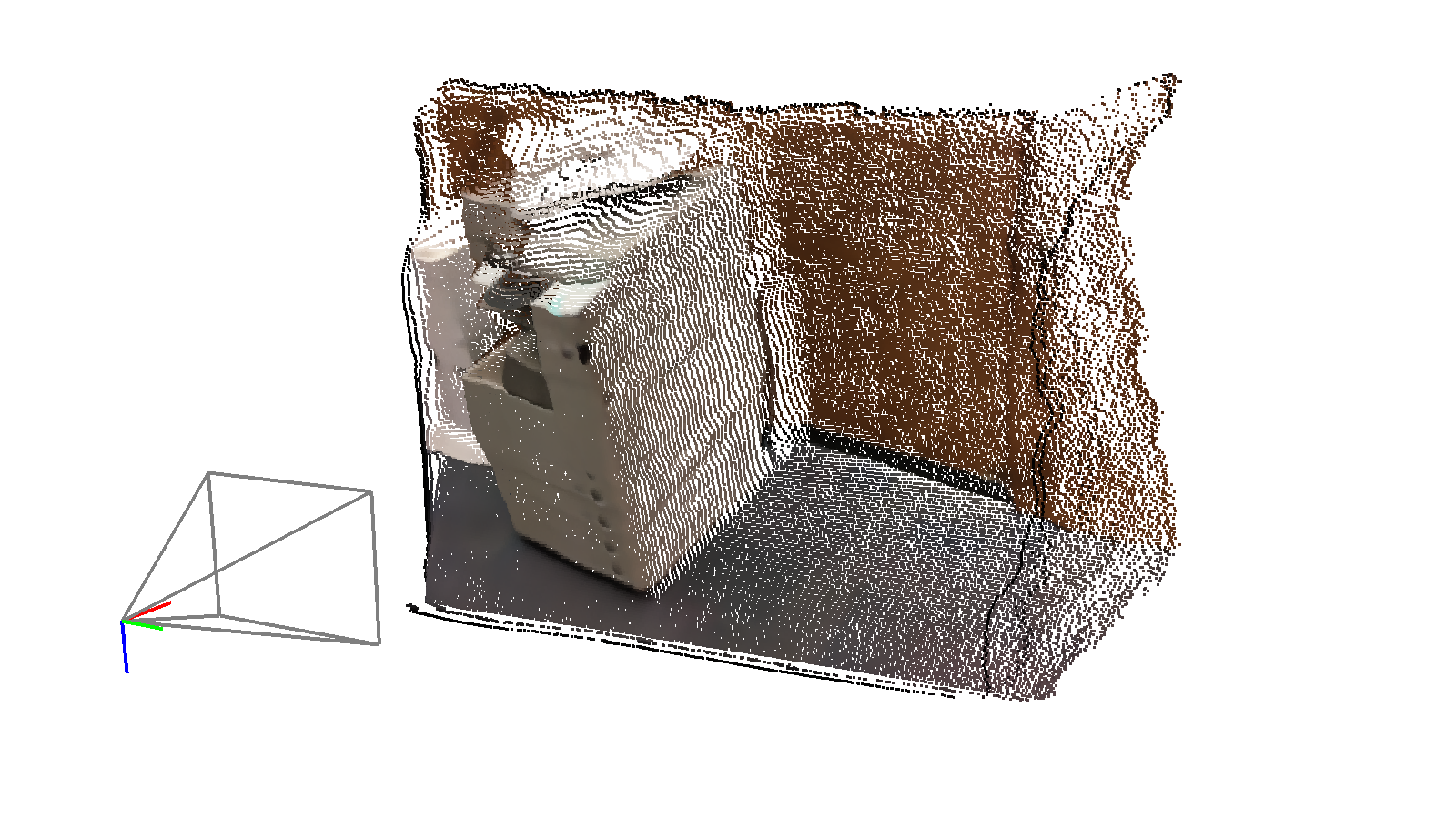}
&
\includegraphics[width=0.2\textwidth
,trim={5.0cm 5.0cm 10.0cm 1.0cm},clip
]{
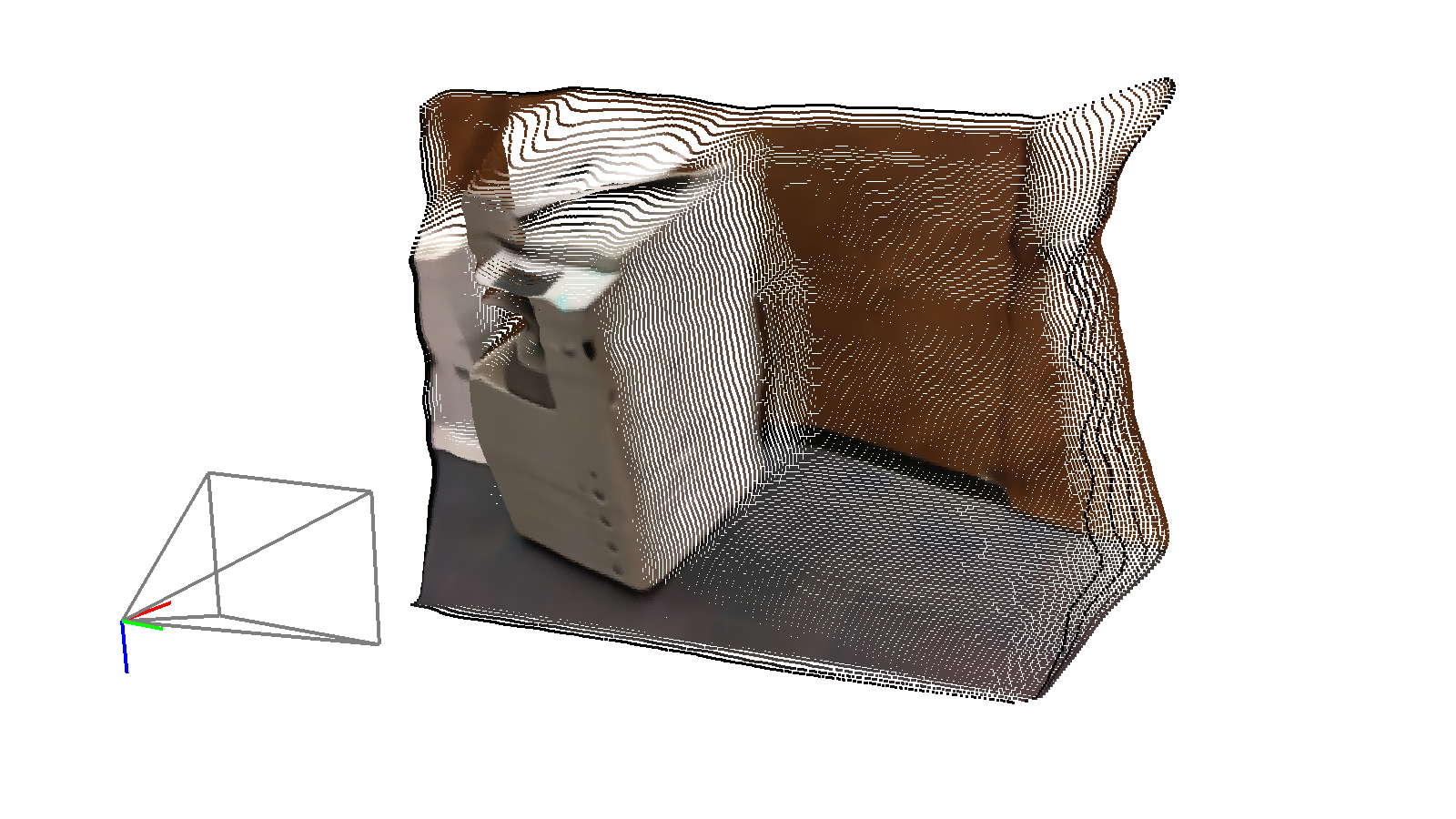}
\end{tabular}
\vspace{-2mm}
\caption{
\textbf{Reconstructed pointclouds} from novel views, using color and depth predictions from different decoders.} 
\label{fig:pointclouds}
\vspace{-5mm}
\end{figure}

Here we analyse the various components and design choices of \Acronym, to evaluate the impact of our contributions in the reported results. Our findings are summarized in \Table \ref{table:nmvs_ablation}, with qualitative results in Figs. \ref{fig:qualitative} and \ref{fig:pointclouds}.

\noindent\textbf{Multi-view Photometric Objective.} Firstly, we ablate the multi-view photometric loss, used as additional regularization to the single-frame view synthesis loss. 
By removing this regularization (\textbf{C}), we observe significantly worse depth results (0.245 vs 0.054 AbsRel), as well as some view synthesis degradation (27.64 vs 28.96 PSNR).  
This is evidence that volumetric rendering is unable to properly learn scene geometry with low viewpoint diversity, and that accurate view synthesis can be obtained with degenerate geometries. 
Alternatively, we trained a self-supervised monocular depth network~\cite{monodepth2} using the same data, and achieved substantially better performance than volumetric rendering (\textbf{A}), however still worse than \Acronym  (0.096 vs 0.054 AbsRel, with qualitative results in the supplementary material). 
These indicate that our hybrid approach improves over any single objective: photometric warping explicitly enforces multi-view consistency, while volumetric rendering implicitly learns 3D geometry.
We also experimented with replacing the multi-view photometric objective with COLMAP supervision (\textbf{B}). As pointed out in other works~\cite{nerfingmvs,dsnerf,depthpriorsnerf}, these predictions are too sparse and noisy to be used without pre-trained priors, leading to worse results (0.134 vs 0.054 AbsRel, 26.17 vs 28.96 PSNR). 

\noindent\textbf{Architecture.} Next, we compare our auto-decoder architecture with a standard NeRF-style MLP~\cite{mildenhall2020nerf} (\textbf{D}). Instead of decoding from the latent space, we map volumetric embeddings $\mathcal{E}_{vol}$ directly into ($\textbf{c},\sigma)$ estimates (\Section \ref{sec:decoders}). As we can see, this approach leads to worse results in depth synthesis (0.068 vs 0.054 AbsRel) and view synthesis (27.50 vs 28.96 PSNR), however it still benefits from photometric regularization (0.068 vs 0.238 AbsRel when removed) (\textbf{E}). This is in accordance with~\cite{rebain2022attention}, in which, for radiance and light field networks, an auto-decoder with a learned latent representation outperformed MLP baselines.

Moreover, replacing our residual light field decoder \cite{wang2022r2l} with a single MLP (\textbf{F}) degrades novel view synthesis (27.34 vs 28.96 PSNR), due to a lack of high frequency details. 

\noindent\textbf{Joint decoding.} Here we consider the joint learning of depth, light, and radiance field predictions from the same latent space.
We evaluate models capable of only volumetric rendering (\textbf{G}), or only light and depth field synthesis (\textbf{H}). 
As expected, depth and light field-only predictions greatly degrade without the view diversity from virtual cameras, that leads to overfitting (0.123 vs 0.054 AbsRel, 23.56 vs 28.96 PSNR). 
Interestingly, volumetric-only predictions also degraded, which we attribute to the use of a shared latent space, that is optimized to store both representations. 

\noindent\textbf{Distillation.}  We also investigated augmenting a pre-trained volumetric representation to also decode depth and light field predictions.
Three scenarios were considered: separate latent spaces (\textbf{I}), and shared latent spaces with (\textbf{J}) and without (\textbf{K}) virtual camera augmentation (VCA). %
When separate latent spaces are used, we observe a substantial improvement in depth and light field performance over single-task learning (0.123 vs 0.063 AbsRel, 27.52 vs 28.96 PSNR). We attribute this behavior to VCA, since this is the only way these two representations interact with each other. 
Interestingly, a similar performance is achieved using shared latent spaces (0.066 vs 0.063 AbsRel, 27.52 vs 27.33 PSNR), even though $\mathcal{S}_R$ is no longer optimized. 
This is an indication that the radiance latent space can be repurposed for depth and light field decoding without further training.  
Moreover, removing VCA in this setting did not degrade performance nearly as much as when separate latent spaces were used (0.076 vs 0.123 AbsRel, 26.16 vs 23.56 PSNR). This is further evidence that radiance representation provides meaningful features for depth and light field decoding, including the preservation of implicit multi-view consistency.

\captionsetup[table]{skip=6pt}
\begin{table}[t!]
\small
\renewcommand{\arraystretch}{0.85}
\centering
{
\setlength{\tabcolsep}{0.3em}
\begin{tabular}{l|c||c|c}
\toprule
\multirow{2}[2]{*}{\textbf{Version}} & 
\multirow{2}[2]{*}{\rotatebox{90}{\scriptsize{Decoder}}} & \multicolumn{2}{c}{Inference} 
\\
\cmidrule(lr){3-4} 
& & \textit{Speed (FPS)} & \textit{Memory (GB)}
\\
\toprule
NeRF & - & 0.35 & 36.54 \\
\midrule
\Acronym (w/o DFG) & \scriptsize{R} & 0.82 & 29.98 \\
\Acronym (1-MLP) & \scriptsize{L} & 351.1 & 4.34 \\
\midrule
\multirow{3}[1]{*}{\Acronym} 
& \scriptsize{D} & 378.4 &  4.21 \\
& \scriptsize{L} & 118.2 &  6.88 \\
& \scriptsize{R} &  37.3 & 11.75 \\
\midrule
\bottomrule
\end{tabular}
}
\caption{
\textbf{Efficiency analysis} for different \Acronym decoders, at inference time (with $192 \times 320$ resolution). \emph{w/o DFG} indicates the removal of depth field guidance, and \emph{1-MLP} indicates a single linear layer in the light field decoder.
}
\label{table:efficiency}
\vspace{-7mm}
\end{table}

\noindent\textbf{Joint training.} Here we evaluate the benefits of jointly training volumetric, depth, and light fields under different conditions. Three settings were considered: the use of separate latent spaces $\mathcal{S}_R$ and $\mathcal{S}_{DL}$ (\textbf{L}), as well as the removal of VCA (\textbf{M}) or depth field guidance (DFG) (\textbf{N}). A key result is that the use of a shared latent space improves results (0.061 vs 0.054 AbsRel, 27.77 vs 28.96 PSNR), as further evidence that both representations produce similar learned features. Moreover, the removal of VCA or DFG leads to overall degradation. Finally, we show that vanishing $\mathcal{L}_p$ (Sec. \ref{sec:schedule}) improves novel view synthesis, by enabling the proper modeling of view-dependent artifacts (\textbf{O}). Interestingly, it does not degrade depth synthesis, indicating that our learned geometry is stable and will not degrade if the photometric regularization is removed. However, it is fundamental in the initial stages of training, as shown in (\textbf{C}) when it is removed altogether (0.245 vs 0.054 AbsRel).

\subsection{Computational Efficiency}
\label{sec:comp_efficiency}

In \Table \ref{table:efficiency} we report inference times and memory requirements using different \Acronym decoders and variations (for hardware details please refer to the supplementary material). Two different components are ablated: depth field guidance (DFG), as described in the \Section \ref{sec:guidance}, and the number of MLP layers in the light field decoder. As expected, depth and light field predictions are substantially faster than volumetric predictions. Furthermore, volumetric prediction efficiency can be greatly improved using DFG ($0.82$ to $37.3$ FPS) to decrease the number of required ray samples (note that this improvement includes the additional cost of evaluating the depth field decoder). Interestingly, even without DFG our auto-decoder architecture is roughly 2 times faster than a traditional NeRF-style MLP~\cite{mildenhall2020nerf}. Moreover, using a single MLP layer for light field decoding speeds up inference by roughly $3$ times ($118.2$ to $378.4$ FPS), at the cost of some degradation in novel view synthesis (\Table \ref{table:nmvs_ablation}, \textbf{F}).

\section{Limitations}

Our method still operates on a scene-specific setting, and thus has to be retrained for new scenes. 
Our method also does not address other traditional limitations of NeRF-like approaches, such as extrapolation to unseen areas and unbounded outdoor scenes. However, our contributions (i.e., the shared latent representation and photometric regularization) can be directly used to augment methods that focus on such scenarios. 
Finally, \Acronym requires overlap between images to enable multi-view photometric self-supervision, and thus is not suitable for very sparse views.

\section{Conclusion}

This paper introduces the multi-view photometric objective as regularization for volume rendering, to mitigate shape-radiance ambiguity and promote the learning of geometrically-consistent representations in cases of low viewpoint diversity. 
To further leverage the geometric properties of this learned latent representation, we propose \Acronym, a novel transformer architecture for the joint learning of depth, light, and radiance fields. 
We show that these three tasks can be encoded into the same shared latent representation, leading to an overall increase in performance over single-task learning without additional network complexity.
As a result, \Acronym establishes a new state of the art in the ScanNet benchmark, outperforming methods that rely on explicit priors from pre-trained depth networks and noisy supervision, while also enabling real-time depth and view synthesis from novel viewpoints.

{\small
\bibliographystyle{ieee_fullname}
\bibliography{reference}
}

\clearpage
\appendix

\begin{center}
\textbf{\large Supplementary Materials}
\end{center}

\section{Dataset Details}
We use ScanNet to evaluate our method. Several splits are popular in recent scene-level NeRF works, hence we consider two popular splits to compare to prior work.

\textbf{ScanNet-Frontal} follows the ScanNet split and evaluation protocol from NerfingMVS~\cite{nerfingmvs}: eight scenes ($0000\_01$, $0079\_00$, $0158\_00$, $0316\_00$, $0521\_00$, $0553\_00$, $0616\_00$, and $0653\_00$) are selected, each with $40$ images covering a local region.
From these, $35$ images are used for training and $5$ are held out for testing.  All images are resized to $484 \times 648$ resolution, and median ground truth scaling is used for depth evaluation. 

\textbf{ScanNet-Rooms} follows the ScanNet split and evaluation protocol from DDP-NeRF~\cite{depthpriorsnerf}: three scenes ($0710\_00$, $0758\_00$, and $0781\_00$) were selected, from which $18$ to $20$ training images and $8$ testing images were extracted.
 All images are resized to $468 \times 624$, and median ground truth scaling is used for depth evaluation. The scenes considered are $0710\_00$, $0758\_00$, and $0781\_00$. To increase frame overlap, such that the multi-view photometric objective has a stronger self-supervised training signal, we included forward and backward context frames for each training image, using a stride of $5$. All other methods were re-evaluated under these new conditions, using officially released open-source repositories and the guidelines described in \cite{depthpriorsnerf}.

\section{Implementation Details}

\subsection{Training parameters}
\label{sec:training}

We implemented our models\footnote{Open-source code and pre-trained weights to replicate our results will be made available upon publication.} using PyTorch~\cite{NEURIPS2019_9015},
with distributed training across eight V100 GPUs. We used grid search to select training parameters, including photometric loss weight $\alpha_p=0.1$, virtual camera loss weight $\alpha_v=0.5$, virtual camera projection noise $\sigma_v=0.25$, depth guidance noise $\sigma_g=0.1$, number of ray samples $K=128$ and depth field guidance samples $K_g=32$, minimum $d_{min}=0.1$ and maximum $d_{max}=5.0$ depth ranges, and a batch size of $b=1$ per GPU.
We use the AdamW optimizer~\cite{loshchilov2019decoupled}, with standard parameters $\beta_1=0.9$, $\beta_2=0.999$, a weight decay of $w=10^{-4}$, and an initial learning rate of $lr=2 \cdot 10^{-4}$. We train for $4000$ epochs, and multiply the learning rate by $0.8$ at each $1000$ epochs. 
A downsample of $4$ is used during training for strided ray sampling, and at test time full resolution estimates are decoded. At each each iteration, $3$ additional images are randomly sampled from the same scene to serve as context. Our self-supervised photometric objective includes auto-masking and minimum reprojection error, as introduced in~\cite{monodepth2}.

\subsection{Architecture Details}

We use $K_o=K_r=K_x=16$ as the number of Fourier frequencies for geometric embeddings (camera center, viewing rays, and sampled 3D points respectively), with maximum resolution $\mu_o = \mu_r = \mu_x = 64$. Our volumetric ($\mathcal{E}_o \oplus \mathcal{E}_r = \mathcal{E}_{vol}$) and ray ($\mathcal{E}_o \oplus \mathcal{E}_x = \mathcal{E}_{ray}$) embeddings both have dimensionality $126 + 126 = 252$. The latent space $\mathcal{S}$ used to encode scene information is of dimensionality $N_l \times D_l = 1024 \times 1024$ (an ablation study regarding this design choice can be found in \Section \ref{sec:latent_space}). Our decoder is composed of a single cross-attention layer, with GeLU as the hidden activation function, dropout of $0.1$, and $2$ attention heads. A single linear layer is then used to project the cross-attention output from $252$ channels to the desired task dimensionality: $O_r=4$ for radiance, $O_l=3$ for light, and $O_d=1$ for depth fields. Alternatively, we experimented with the deep residual network of~\cite{wang2022r2l} as the decoder, achieving significant improvements in light field novel view synthesis at the expense of slower inference times (\Table 5 in the main paper).

\begin{figure*}[t!]
\centering
\subfloat{
\includegraphics[width=0.48\textwidth,trim={1.1cm 0.5cm 0.5cm 0.4cm},clip]{
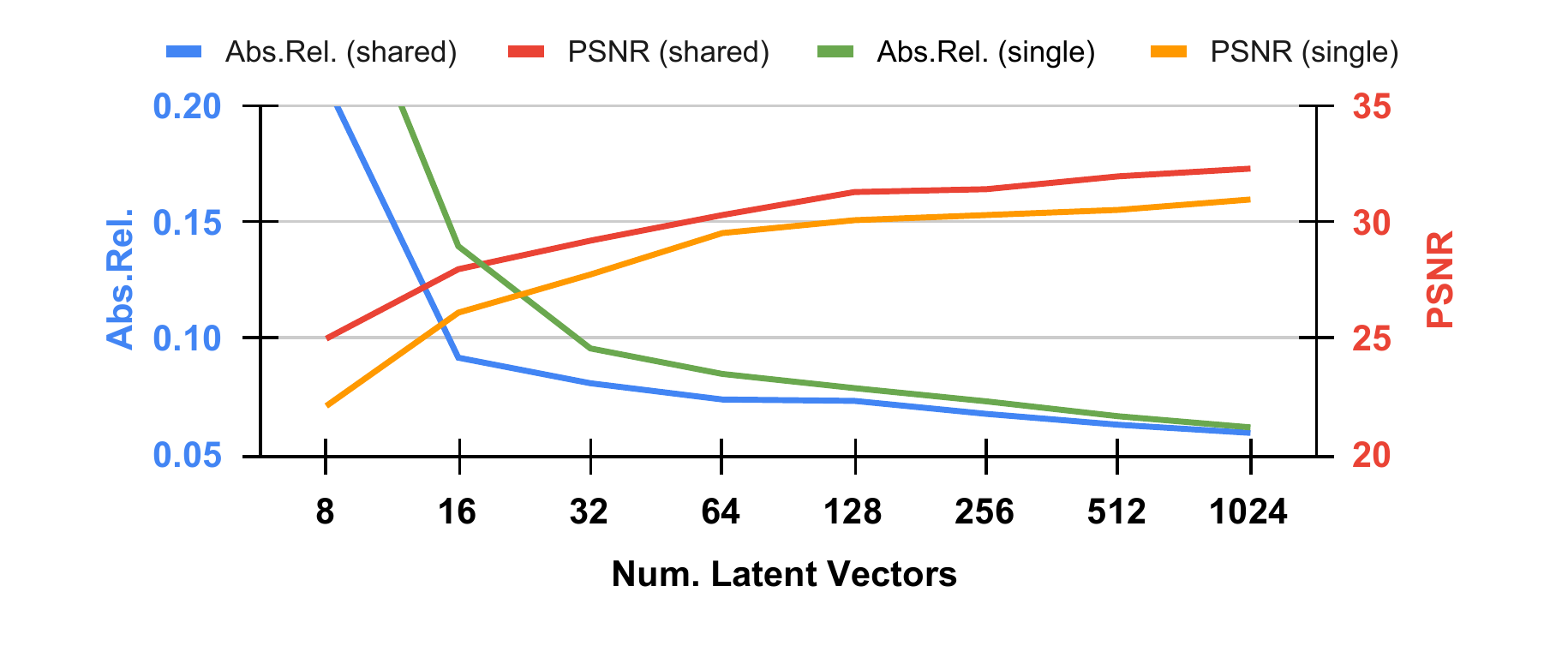}}
\hspace{3mm}
\subfloat{
\includegraphics[width=0.48\textwidth,trim={1.1cm 0.5cm 0.5cm 0.4cm},clip]{
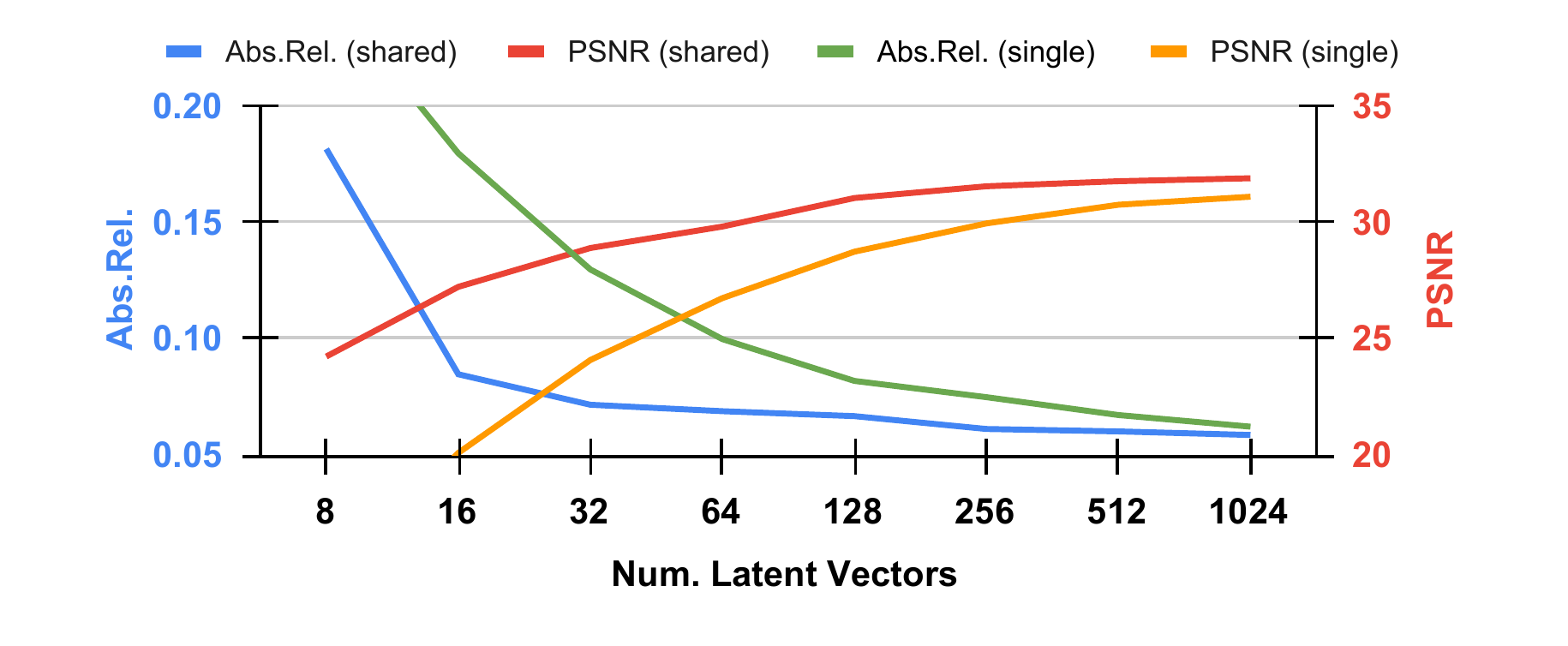}}
\\
\vspace{3mm}
\setcounter{subfigure}{0}
\subfloat[Volumetric rendering predictions]{
\raisebox{2mm}{\includegraphics[width=0.48\textwidth,trim={1.1cm 0.5cm 0.5cm 0.4cm},clip]{
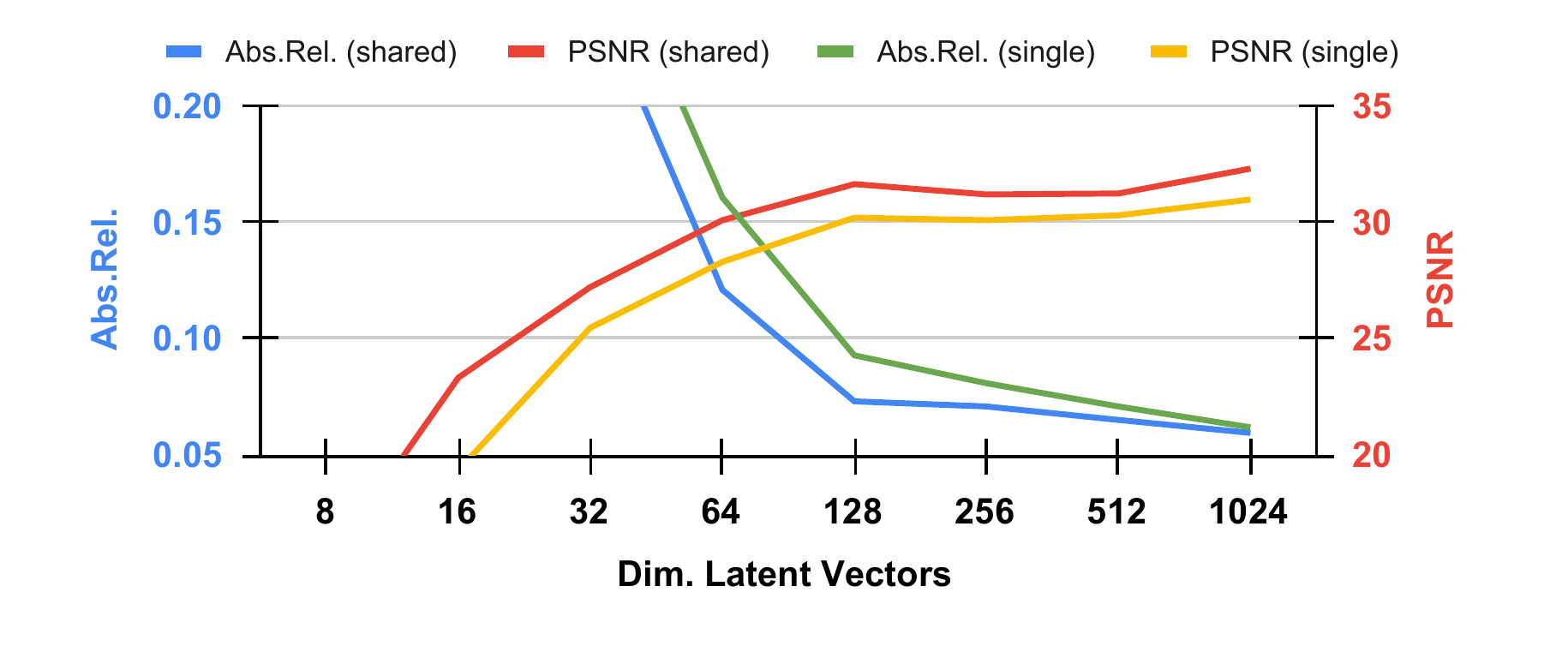}}}
\hspace{3mm}
\subfloat[Depth and light field predictions] {
\raisebox{2mm}{\includegraphics[width=0.48\textwidth,trim={1.1cm 0.5cm 0.5cm 0.4cm},clip]{
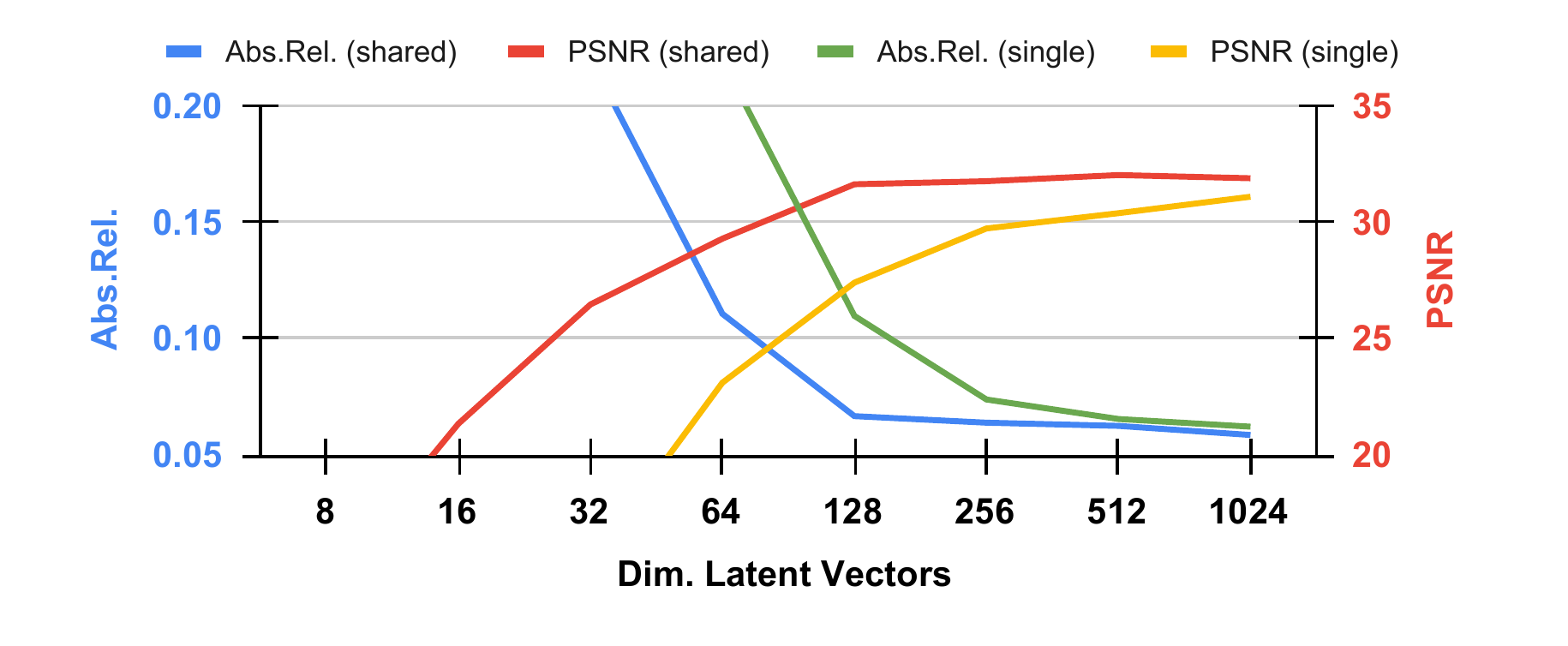}}}
\caption{
\textbf{Depth and view synthesis performance} on \emph{ScanNet} (scene $0653\_00$), with varying latent space shapes (larger values were not considered due to computational constraints). Blue and red lines correspond to predictions decoded from a shared latent space, and green and yellow lines to predictions decoded from latent spaces with a single representation. We observe that sharing the latent space between representations not only does not degrade results, but in fact leads to overall improvements in both view synthesis and depth estimation. These improvements are more noticeable in smaller latent spaces, particularly for depth and light field estimates, indicating that both representations are compatible for multi-task decoding.
}
\label{fig:smaller2}
\end{figure*}

\section{Latent Space Dimensionality}
\label{sec:latent_space}

Here we analyze the impact that changing the dimensions of the latent space $\mathcal{S}$ has on performance, both in terms of view synthesis (PSNR) and depth estimation (Abs.Rel.). Two variables are considered: the number $N_l$ of latent vectors, and the dimensionality $D_l$ of these vectors. The results of this analysis are shown in \Figure \ref{fig:smaller2} (blue and red lines), where we can see that larger latent spaces indeed leads to an improvement in performance (i.e. better view synthesis PSNR and absolute relative depth error), albeit with diminishing returns. To achieve optimal results without excessive computational cost, in all experiments we used a $1024 \times 1024$ latent space. When experimenting with smaller dimensionalities, we noticed a gradual decrease in performance, followed by a steep change around $N_l=16$ and $D_l=128$. This sudden ``phase transition" indicates the point at which the latent space becomes unable to properly encode the scene representation.

To further evaluate the properties of our learned implicit representation, we performed similar experiments in which two latent spaces are optimized, one containing only a volumetric representation, and another only a light and depth field representation. For a fair comparison, both latent spaces are still trained jointly (i.e., light and depth predictions benefit from virtual volumetric supervision, and volumetric predictions benefit from depth field guidance). The green and yellow lines in \Figure \ref{fig:smaller2} show results using this setting. Interestingly, we observe that maintaining separate latent spaces for each representation not only leads to worse performance than using a single latent space (as we show in the main paper), but also that this performance gap increases when smaller latent spaces are used.  

This is particularly noticeable in the case of depth and light field predictions, that experience the ``phase transition" at significantly higher dimensionalities: $128 \times 256$, compared to $16 \times 128$ when using a shared latent space. We attribute this behavior to the regularization effect that the volumetric representation has on light and depth field predictions. As we show in the main paper (\Section 4.4.2), jointly learning a volumetric representation has a similar effect to virtual camera augmentation, promoting the learning of a multi-view consistent representation for light and depth field predictions. With smaller model complexities, this multi-view consistency becomes a key factor in the learning of a useful representation for accurate predictions from novel viewpoints.

\begin{figure}[t!] 
  \begin{tabular}{cc} 
  \setlength{\tabcolsep}{1pt}
  \multirow{2}{*}{
   \raisebox{-0.5cm}[0cm][0cm]{
  \hspace{-5mm}
  \rotatebox{90}{\scriptsize{NerfingMVS~\cite{nerfingmvs}}}}
  \hspace{-2mm}
   \raisebox{-1.7cm}[0cm][0cm]{
   \includegraphics[width=0.25\textwidth,trim=23cm 0cm 0cm 2cm,clip]{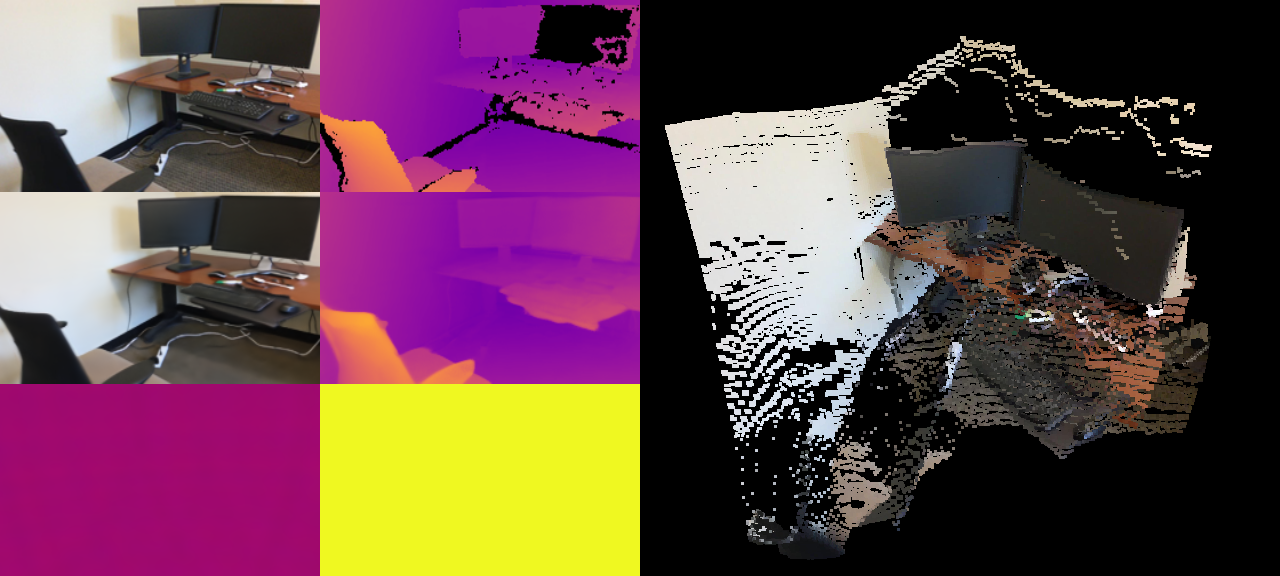}}} \hspace{-4mm}
  & \includegraphics[width=0.19\textwidth]{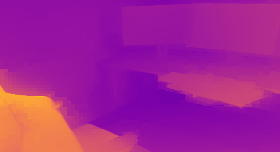} \\
  & \includegraphics[width=0.19\textwidth]{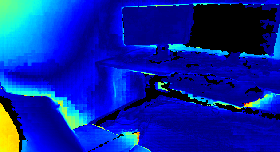} \\
  \multirow{2}{*}{
  \hspace{-5mm}
  \hspace{1mm}
  \rotatebox{90}{\scriptsize{DeLiRa}}
  \hspace{-2mm}
   \raisebox{-1.4cm}[0cm][0cm]{
   \includegraphics[width=0.25\textwidth,trim=23cm 0cm 0cm 2cm,clip]{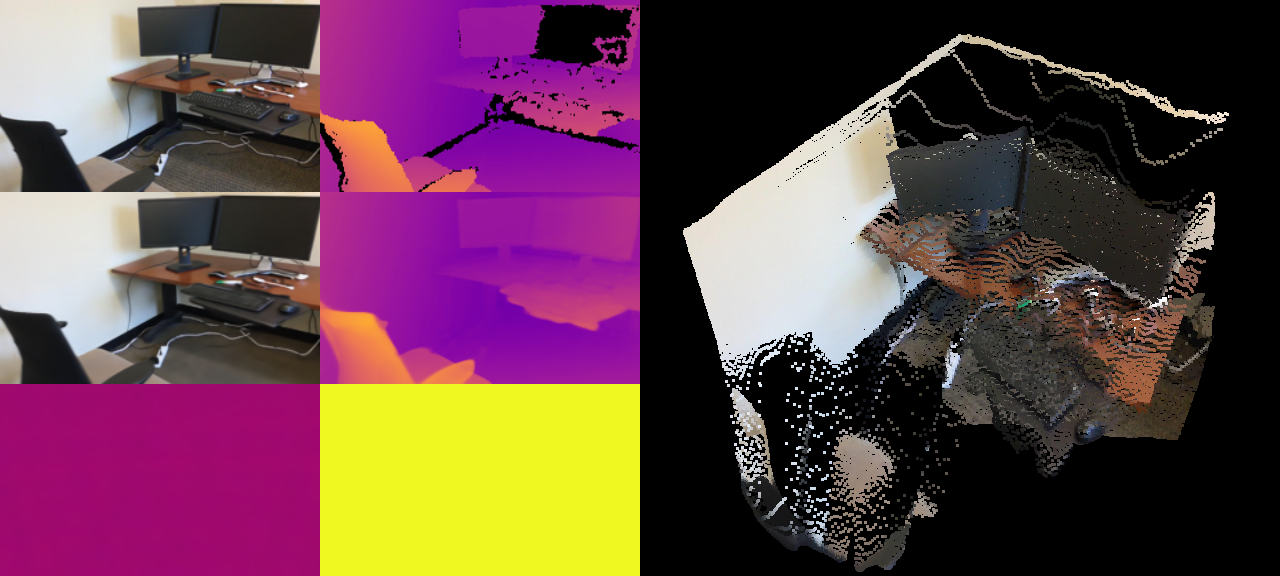}}} \hspace{-4mm}
  & \includegraphics[width=0.19\textwidth]{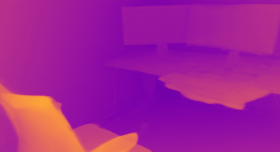} \\
  & \includegraphics[width=0.19\textwidth]{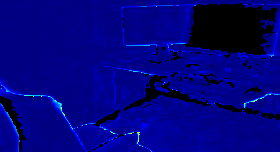} \\
  \end{tabular}
\caption{\textbf{Qualitative comparison between DeLiRa and NerfingMVS~\cite{nerfingmvs}}, for depth estimation from novel viewpoints. We show predicted depth maps (top right), depth error maps (bottom right), and reconstructed pointclouds using predicted depth and colors (left). Our approach leads to sharper depth maps, with errors concentrated on discontinuities around object boundaries, as well as better reconstruction of planar surfaces.
}
\label{fig:comparison}
\vspace{-6mm}
\end{figure}

\section{Additional Qualitative Results}
\label{sec:supp_qualitative_results}
We also include additional qualitative results to complement the ones provided in the main paper. In \Figure \ref{fig:comparison} we compare depth estimation results from \Acronym and those produced by NerfingMVS~\cite{nerfingmvs}, the previous state of the art in \emph{ScanNet-Frontal}. As we can see, our predictions are sharper, with errors concentrated in discontinuities around object boundaries. \Acronym also improves upon NerfingMVS in terms of reconstructing planar surfaces, such as the left wall and the right computer monitor. These improvements are particularly meaningful given that NerfingMVS (and most other current approaches) rely on depth priors from pre-trained networks, while \Acronym is trained using only information from the observed scene. 

In \Figure \ref{fig:qualitative_supp} we show predicted RGB images and depth maps obtained using different \Acronym decoders (cf. Fig. 3 in the main paper). We also provide error maps for both predictions, in the form of normalized absolute differences. As a baseline, we show results produced by a model trained without our contributions (i.e., the multi-view photometric objective and the joint learning of depth, light, and radiance fields). Interestingly, this baseline model achieves novel view synthesis results comparable to our proposed architecture, however depth estimates are considerably worse. These are examples of the shape-radiance ambiguity, in which accurate novel view synthesis can still be achieved even with degenerated learned geometries, especially in cases of limited viewpoint diversity. By introducing the multi-view photometric objective as additional regularization, we promote convergence to the proper scene geometry, improving depth estimation and, by extension, novel view synthesis. Furthermore, our learned latent representation can be queried both in the form of volumetric renderings, via the radiance field decoder, as well as direct depth color estimates, via the depth and light field decoders. 

\begin{figure}[t]
\centering
\subfloat[RGB]{
\includegraphics[width=0.15\textwidth]{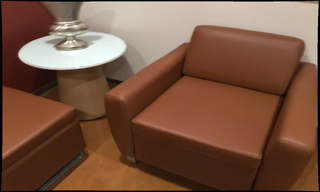}}
\subfloat[Monodepth2]{
\includegraphics[width=0.15\textwidth]{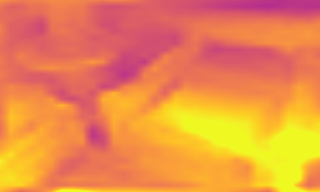}}
\subfloat[DeLiRa]{
\includegraphics[width=0.15\textwidth]{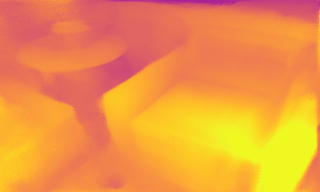}}
\vspace{-2mm}
\caption{\textbf{Qualitative example of depth predictions} between \Acronym and a traditional monocular depth network.}
\label{fig:monodepth}
\vspace{-5mm}
\end{figure}

\begin{figure*}[t!] 
  \begin{tabular}{cccc} 
  \small{Ground Truth} & \small{Baseline (radiance field)} & \small{DeLiRa (radiance field)} & \small{DeLiRa (depth and light fields)} \\
  \cmidrule(lr){1-1} \cmidrule(lr){2-2} \cmidrule(lr){3-3} \cmidrule(lr){4-4}
  \includegraphics[width=0.13\textwidth,height=1.65cm]{images/qualitative_error/B/gtB.png} &
  \includegraphics[width=0.13\textwidth,height=1.65cm]{images/qualitative_error/B/pred1B_base.png}
  \includegraphics[width=0.13\textwidth,height=1.65cm]{images/qualitative_error/B/err1b_base.png} &
  \includegraphics[width=0.13\textwidth,height=1.65cm]{images/qualitative_error/B/pred1B.png}
  \includegraphics[width=0.13\textwidth,height=1.65cm]{images/qualitative_error/B/err1b.png} &
  \includegraphics[width=0.13\textwidth,height=1.65cm]{images/qualitative_error/B/pred2B.png}
  \includegraphics[width=0.13\textwidth,height=1.65cm]{images/qualitative_error/B/err2b.png} \\
  \includegraphics[width=0.13\textwidth,height=1.65cm]{images/qualitative_error/B/gtA.png} &
  \includegraphics[width=0.13\textwidth,height=1.65cm]{images/qualitative_error/B/pred1A_base.png}
  \includegraphics[width=0.13\textwidth,height=1.65cm]{images/qualitative_error/B/err1a_base.png} &
  \includegraphics[width=0.13\textwidth,height=1.65cm]{images/qualitative_error/B/pred1A.png}
  \includegraphics[width=0.13\textwidth,height=1.65cm]{images/qualitative_error/B/err1a.png} &
  \includegraphics[width=0.13\textwidth,height=1.65cm]{images/qualitative_error/B/pred2A.png}
  \includegraphics[width=0.13\textwidth,height=1.65cm]{images/qualitative_error/B/err2a.png} 
  \vspace{3mm} \\ 
  \includegraphics[width=0.13\textwidth,height=1.65cm]{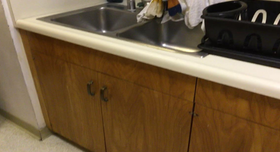} &
  \includegraphics[width=0.13\textwidth,height=1.65cm]{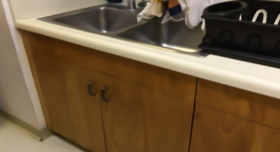}
  \includegraphics[width=0.13\textwidth,height=1.65cm]{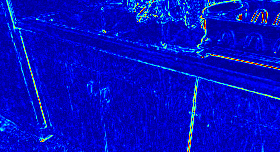} &
  \includegraphics[width=0.13\textwidth,height=1.65cm]{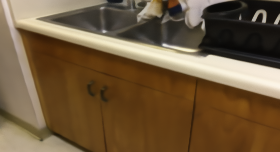}
  \includegraphics[width=0.13\textwidth,height=1.65cm]{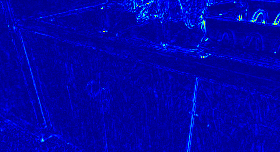} &
  \includegraphics[width=0.13\textwidth,height=1.65cm]{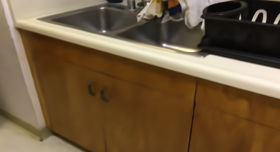}
  \includegraphics[width=0.13\textwidth,height=1.65cm]{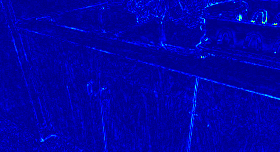} \\
  \includegraphics[width=0.13\textwidth,height=1.65cm]{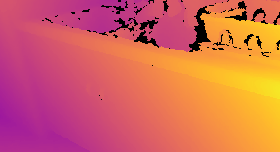} &
  \includegraphics[width=0.13\textwidth,height=1.65cm]{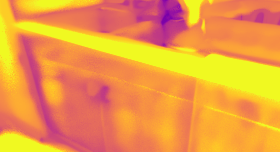}
  \includegraphics[width=0.13\textwidth,height=1.65cm]{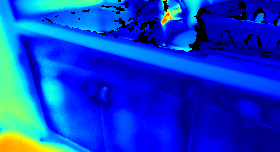} &
  \includegraphics[width=0.13\textwidth,height=1.65cm]{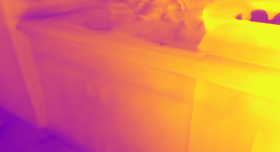}
  \includegraphics[width=0.13\textwidth,height=1.65cm]{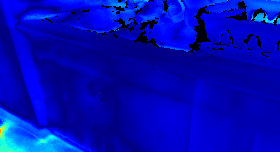} &
  \includegraphics[width=0.13\textwidth,height=1.65cm]{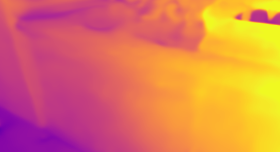}
  \includegraphics[width=0.13\textwidth,height=1.65cm]{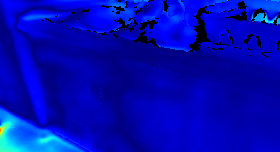} 
  \vspace{3mm} \\ 
  \includegraphics[width=0.13\textwidth,height=1.65cm]{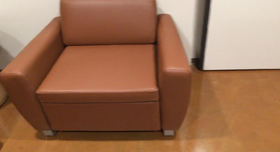} &
  \includegraphics[width=0.13\textwidth,height=1.65cm]{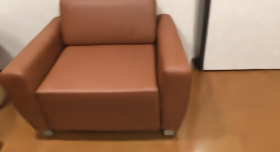}
  \includegraphics[width=0.13\textwidth,height=1.65cm]{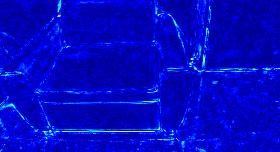} &
  \includegraphics[width=0.13\textwidth,height=1.65cm]{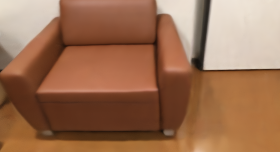}
  \includegraphics[width=0.13\textwidth,height=1.65cm]{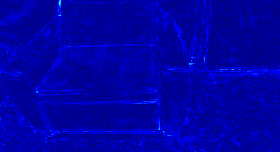} &
  \includegraphics[width=0.13\textwidth,height=1.65cm]{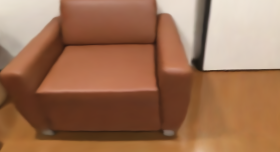}
  \includegraphics[width=0.13\textwidth,height=1.65cm]{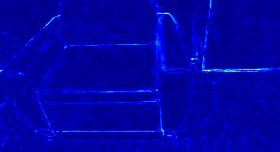} \\
  \includegraphics[width=0.13\textwidth,height=1.65cm]{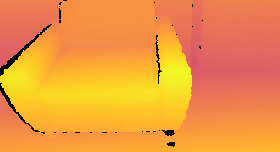} &
  \includegraphics[width=0.13\textwidth,height=1.65cm]{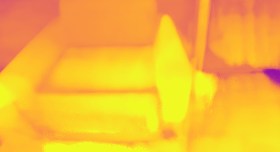}
  \includegraphics[width=0.13\textwidth,height=1.65cm]{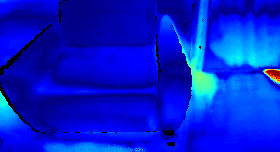} &
  \includegraphics[width=0.13\textwidth,height=1.65cm]{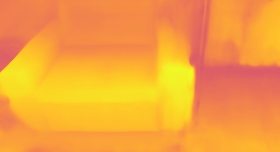}
  \includegraphics[width=0.13\textwidth,height=1.65cm]{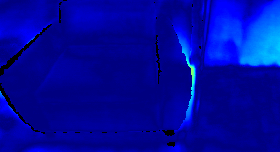} &
  \includegraphics[width=0.13\textwidth,height=1.65cm]{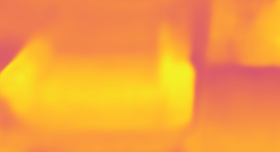}
  \includegraphics[width=0.13\textwidth,height=1.65cm]{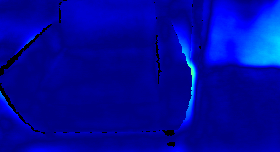} 
  \vspace{3mm} \\ 
  \includegraphics[width=0.13\textwidth,height=1.65cm]{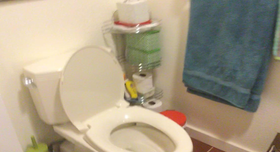} &
  \includegraphics[width=0.13\textwidth,height=1.65cm]{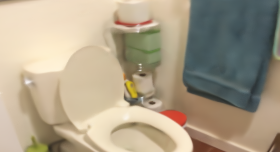}
  \includegraphics[width=0.13\textwidth,height=1.65cm]{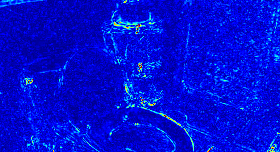} &
  \includegraphics[width=0.13\textwidth,height=1.65cm]{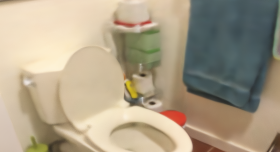}
  \includegraphics[width=0.13\textwidth,height=1.65cm]{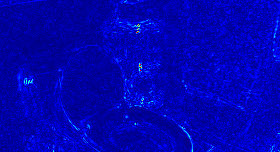} &
  \includegraphics[width=0.13\textwidth,height=1.65cm]{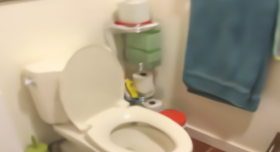}
  \includegraphics[width=0.13\textwidth,height=1.65cm]{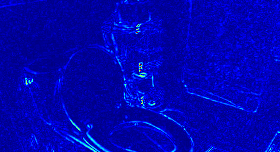} \\
  \includegraphics[width=0.13\textwidth,height=1.65cm]{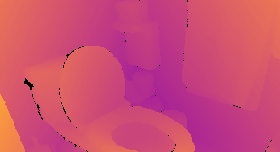} &
  \includegraphics[width=0.13\textwidth,height=1.65cm]{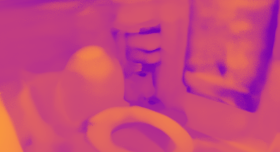}
  \includegraphics[width=0.13\textwidth,height=1.65cm]{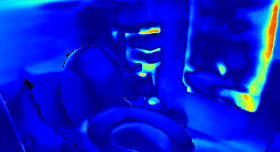} &
  \includegraphics[width=0.13\textwidth,height=1.65cm]{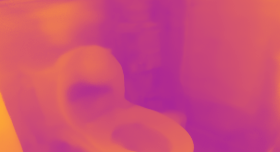}
  \includegraphics[width=0.13\textwidth,height=1.65cm]{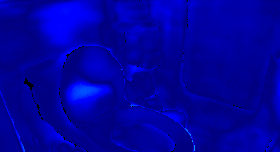} &
  \includegraphics[width=0.13\textwidth,height=1.65cm]{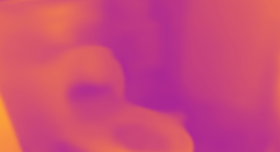}
  \includegraphics[width=0.13\textwidth,height=1.65cm]{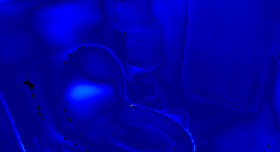}  
  \vspace{3mm} \\ 
  \includegraphics[width=0.13\textwidth,height=1.65cm]{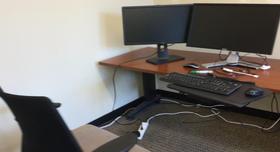} &
  \includegraphics[width=0.13\textwidth,height=1.65cm]{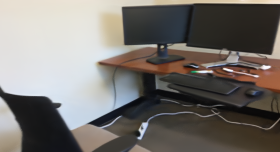}
  \includegraphics[width=0.13\textwidth,height=1.65cm]{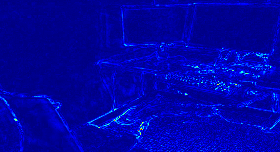} &
  \includegraphics[width=0.13\textwidth,height=1.65cm]{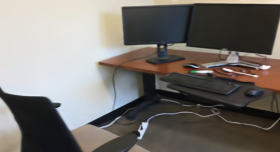}
  \includegraphics[width=0.13\textwidth,height=1.65cm]{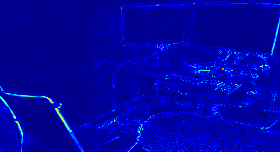} &
  \includegraphics[width=0.13\textwidth,height=1.65cm]{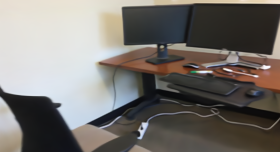}
  \includegraphics[width=0.13\textwidth,height=1.65cm]{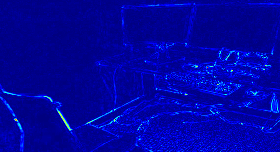} \\
  \includegraphics[width=0.13\textwidth,height=1.65cm]{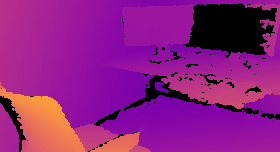} &
  \includegraphics[width=0.13\textwidth,height=1.65cm]{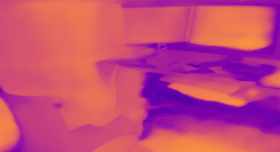}
  \includegraphics[width=0.13\textwidth,height=1.65cm]{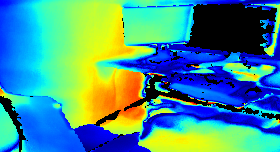} &
  \includegraphics[width=0.13\textwidth,height=1.65cm]{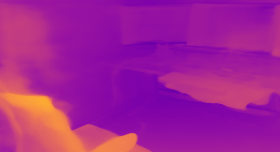}
  \includegraphics[width=0.13\textwidth,height=1.65cm]{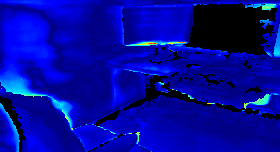} &
  \includegraphics[width=0.13\textwidth,height=1.65cm]{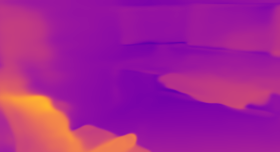}
  \includegraphics[width=0.13\textwidth,height=1.65cm]{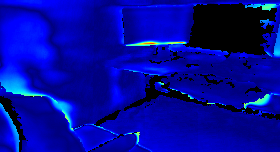}     
  \end{tabular}
\caption{
\textbf{Additional qualitative depth and view synthesis results} from unseen viewpoints, using different \Acronym decoders. As a baseline, we show predictions obtained from a model trained without our contributions, leading to a degenerate learned geometry due to shape-radiance ambiguity (i.e., accurate view synthesis with poor depth predictions).
RGB and depth error maps are calculated as absolute differences and respectively normalized between $[0.0,0.5]$ and $[0.0,1.0]$. 
}
\label{fig:qualitative_supp}
\end{figure*}

\begin{figure*}[t!] 
  \begin{tabular}{ccc} 
  \small{Ground Truth} & \small{DeLiRa (radiance field)} & \small{DeLiRa (depth and light fields)} \\
  \cmidrule(lr){1-1} \cmidrule(lr){2-2} \cmidrule(lr){3-3}
  \includegraphics[width=0.3\textwidth,trim=23cm 0cm 0cm 2cm,clip]{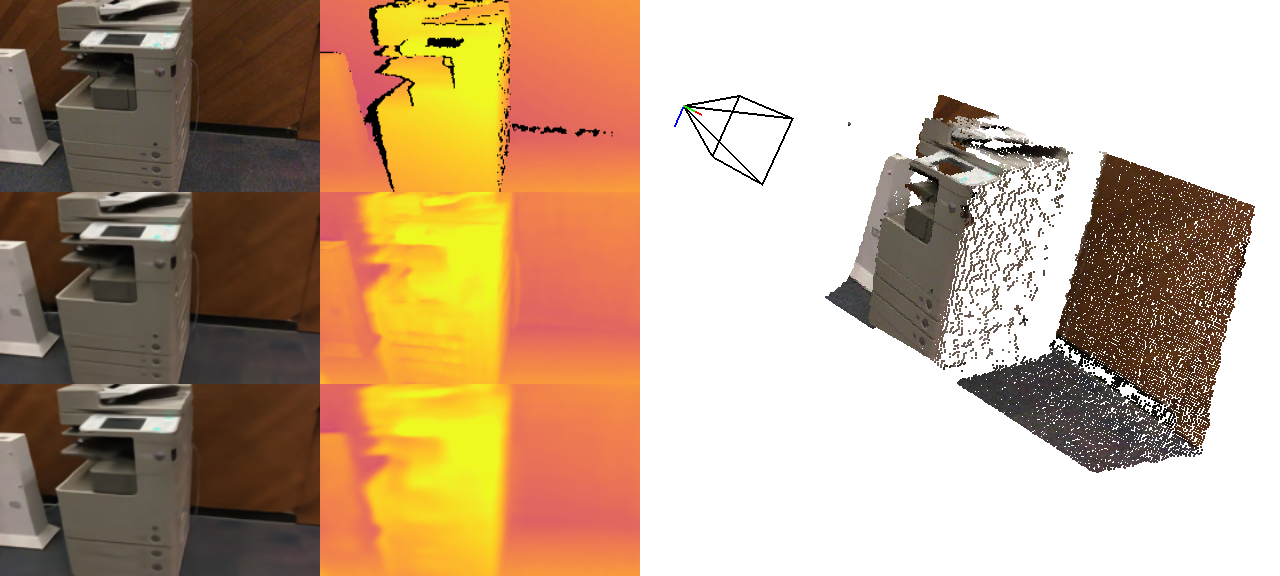} &
  \includegraphics[width=0.3\textwidth,trim=23cm 0cm 0cm 2cm,clip]{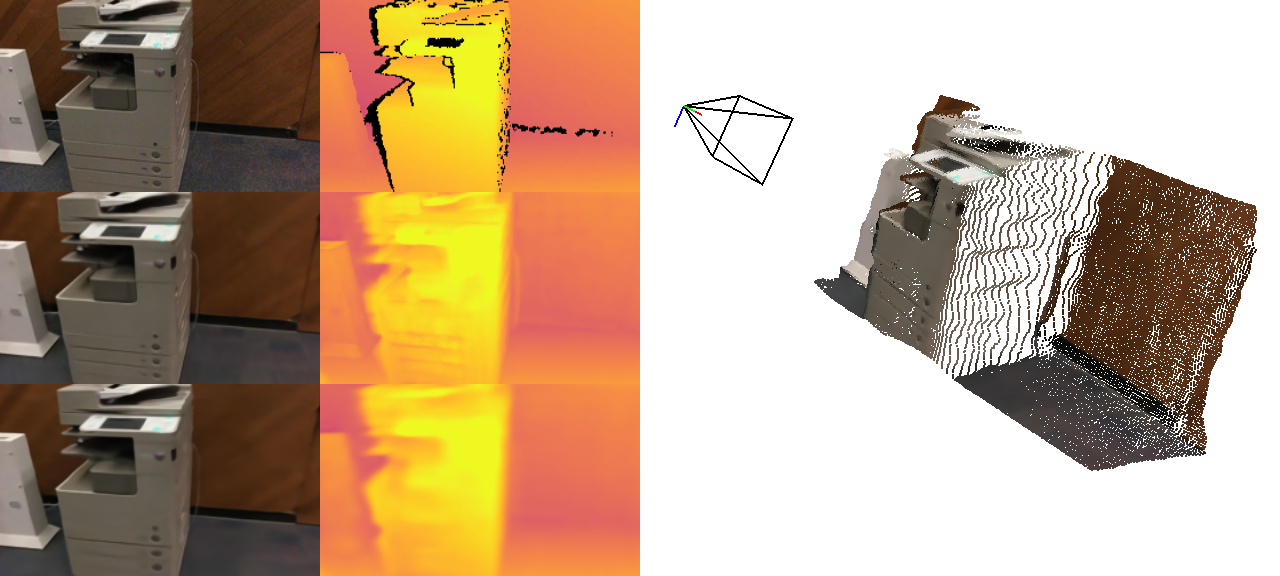} &
  \includegraphics[width=0.3\textwidth,trim=23cm 0cm 0cm 2cm,clip]{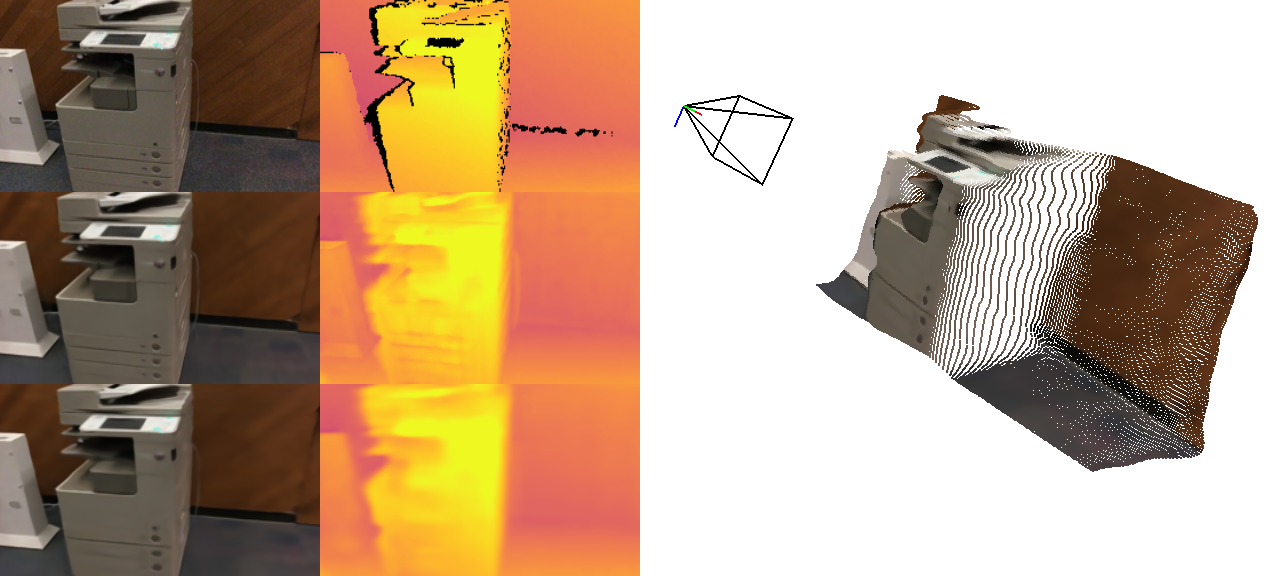}
  \vspace{-5mm} \\ 
  \includegraphics[width=0.3\textwidth,trim=23cm 0cm 0cm 2cm,clip]{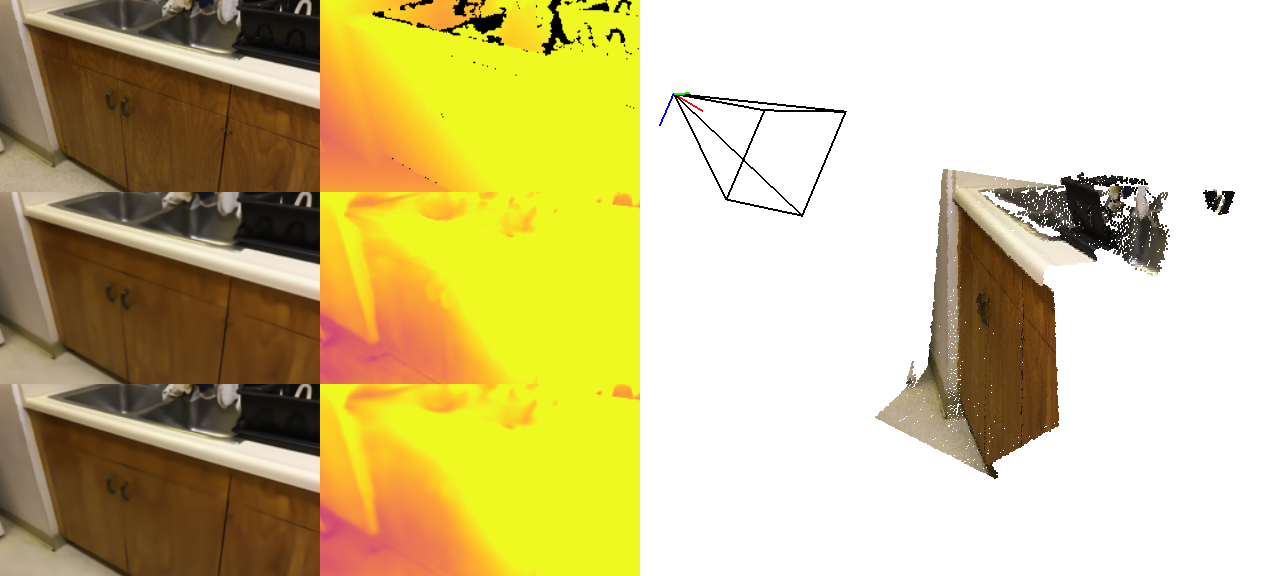} &
  \includegraphics[width=0.3\textwidth,trim=23cm 0cm 0cm 2cm,clip]{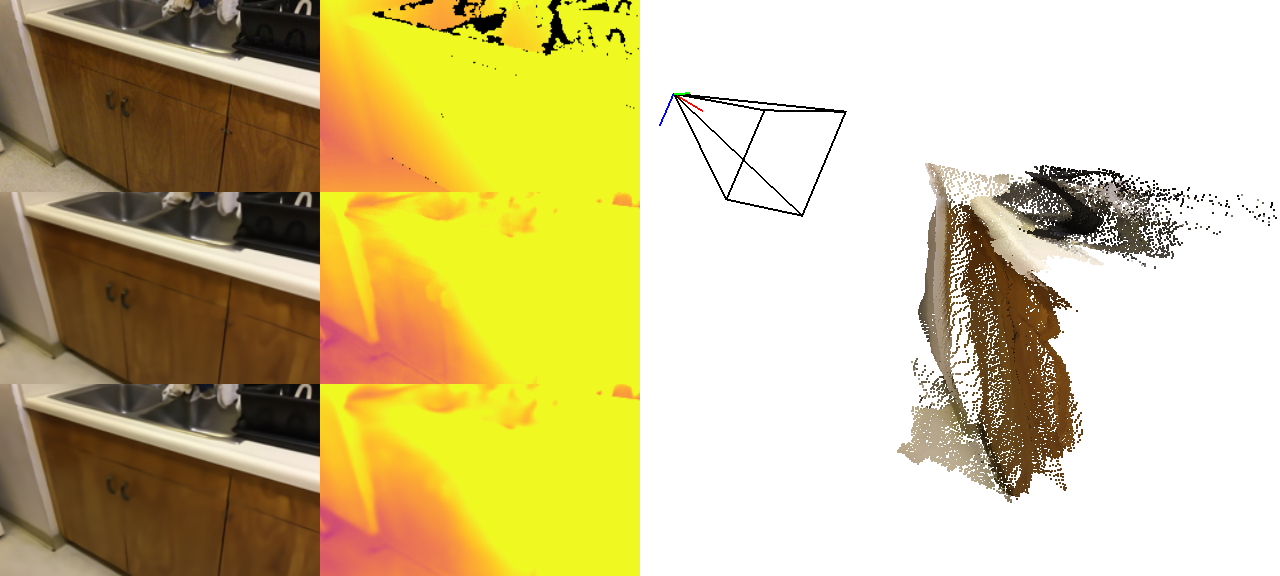} &
  \includegraphics[width=0.3\textwidth,trim=23cm 0cm 0cm 2cm,clip]{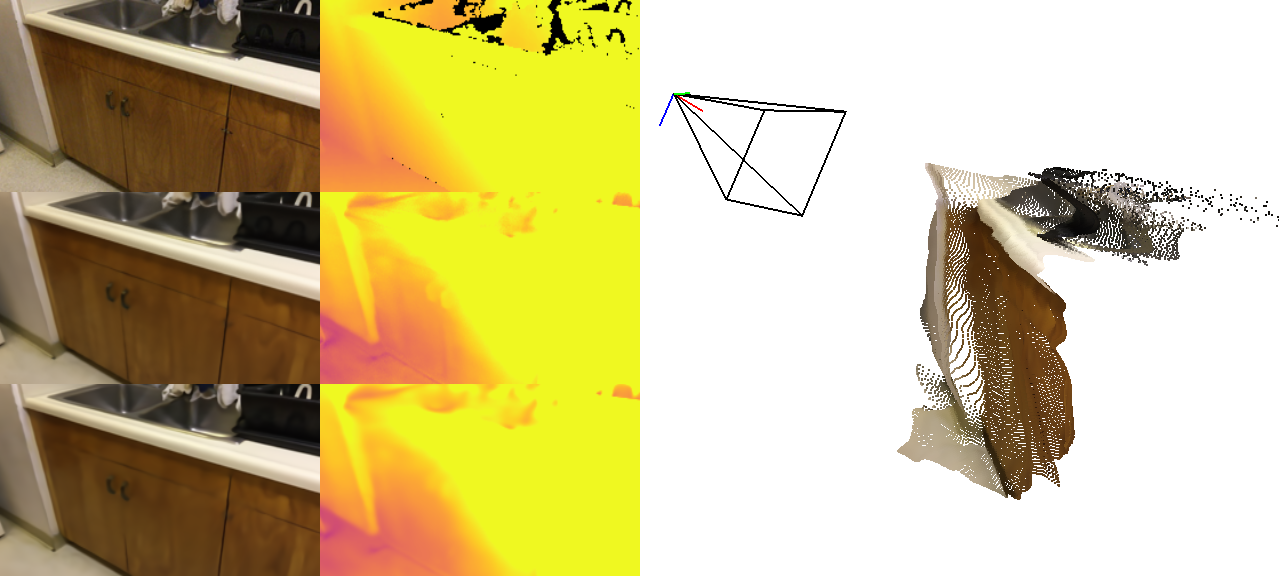}
  \vspace{-5mm} \\ 
  \includegraphics[width=0.3\textwidth,trim=23cm 0cm 0cm 2cm,clip]{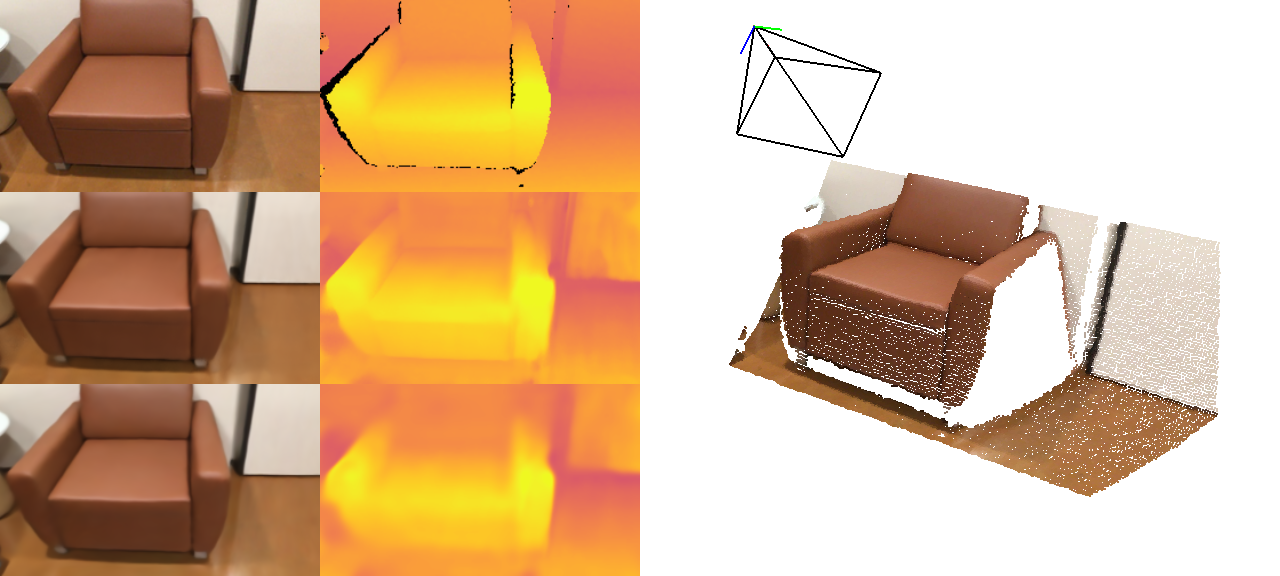} &
  \includegraphics[width=0.3\textwidth,trim=23cm 0cm 0cm 2cm,clip]{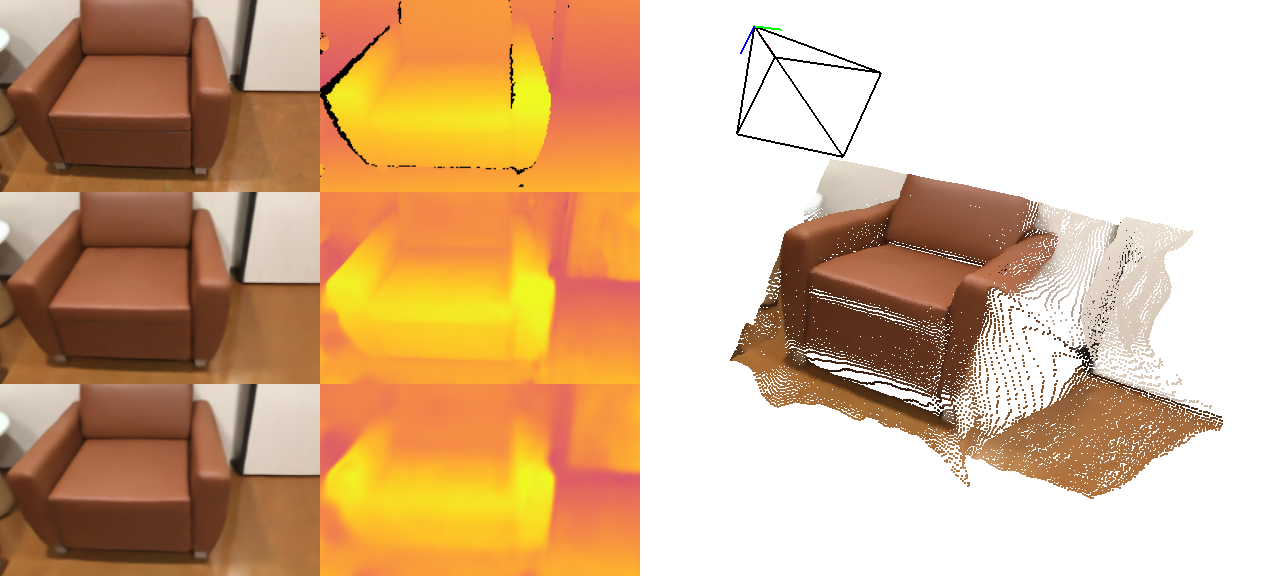} &
  \includegraphics[width=0.3\textwidth,trim=23cm 0cm 0cm 2cm,clip]{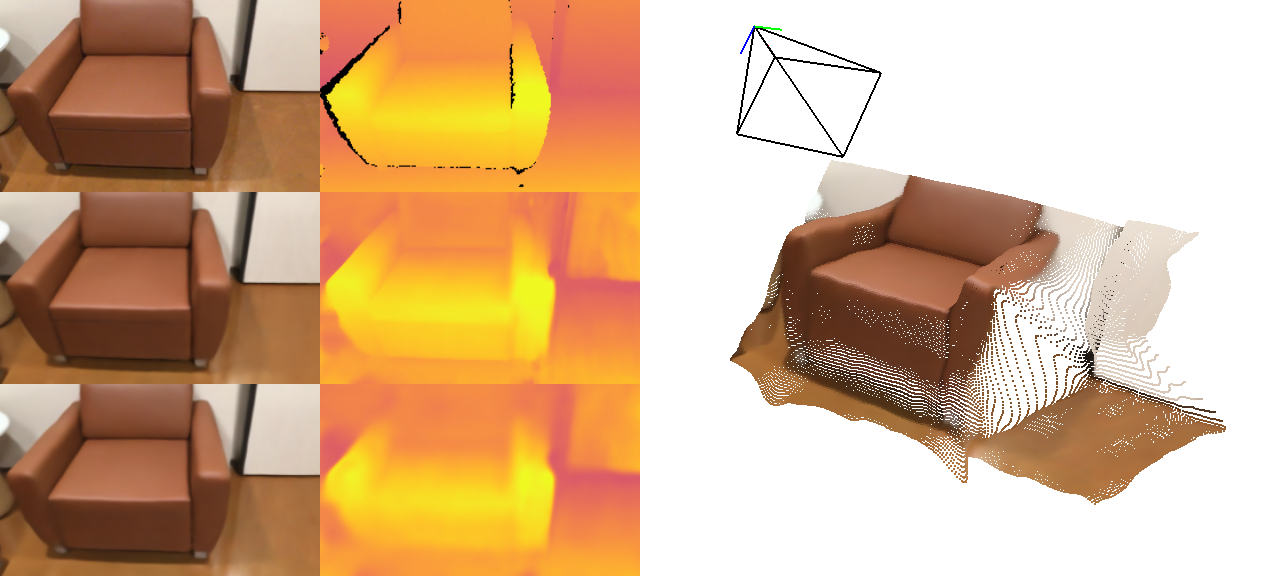}
  \vspace{-5mm} \\ 
  \includegraphics[width=0.3\textwidth,trim=23cm 0cm 0cm 2cm,clip]{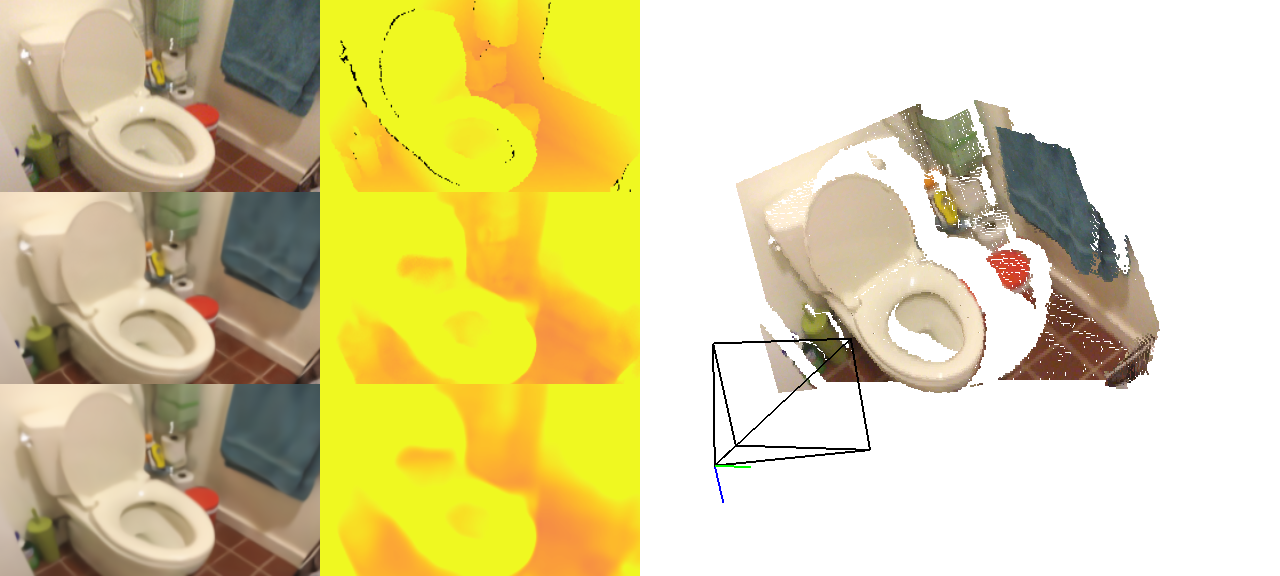} &
  \includegraphics[width=0.3\textwidth,trim=23cm 0cm 0cm 2cm,clip]{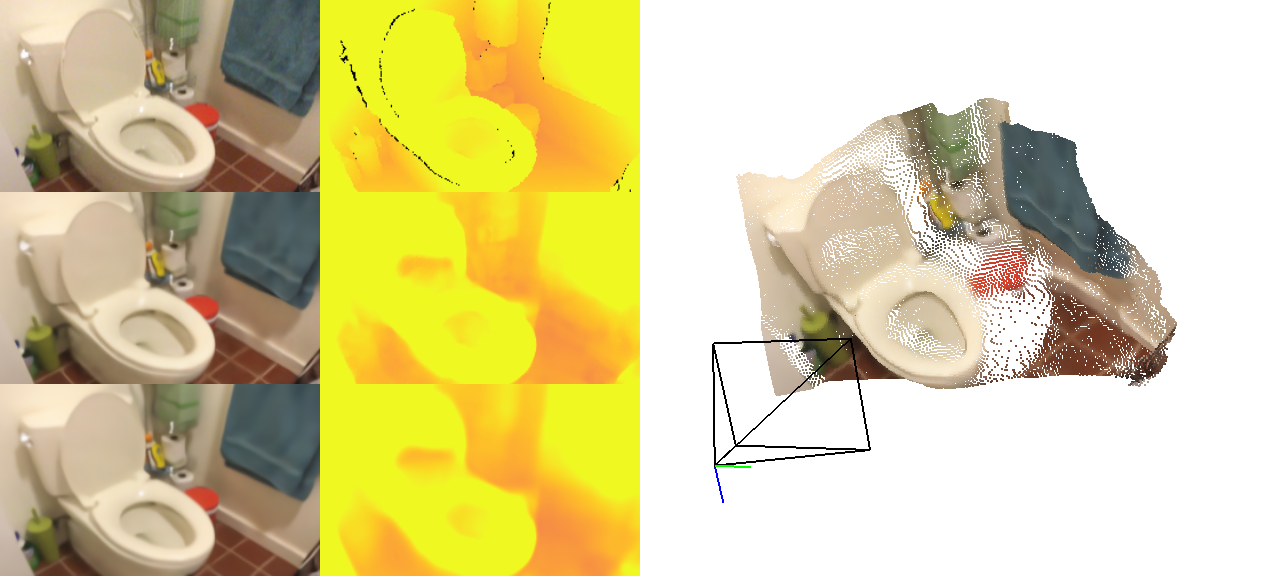} &
  \includegraphics[width=0.3\textwidth,trim=23cm 0cm 0cm 2cm,clip]{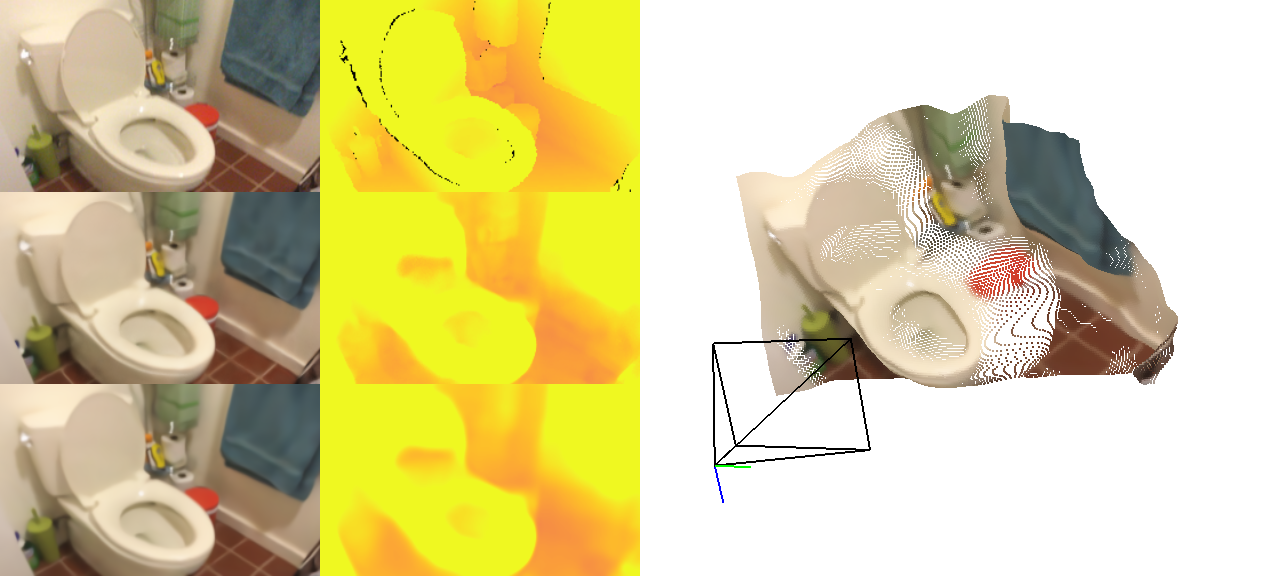}
  \vspace{-5mm} \\ 
  \includegraphics[width=0.3\textwidth,trim=23cm 0cm 0cm 2cm,clip]{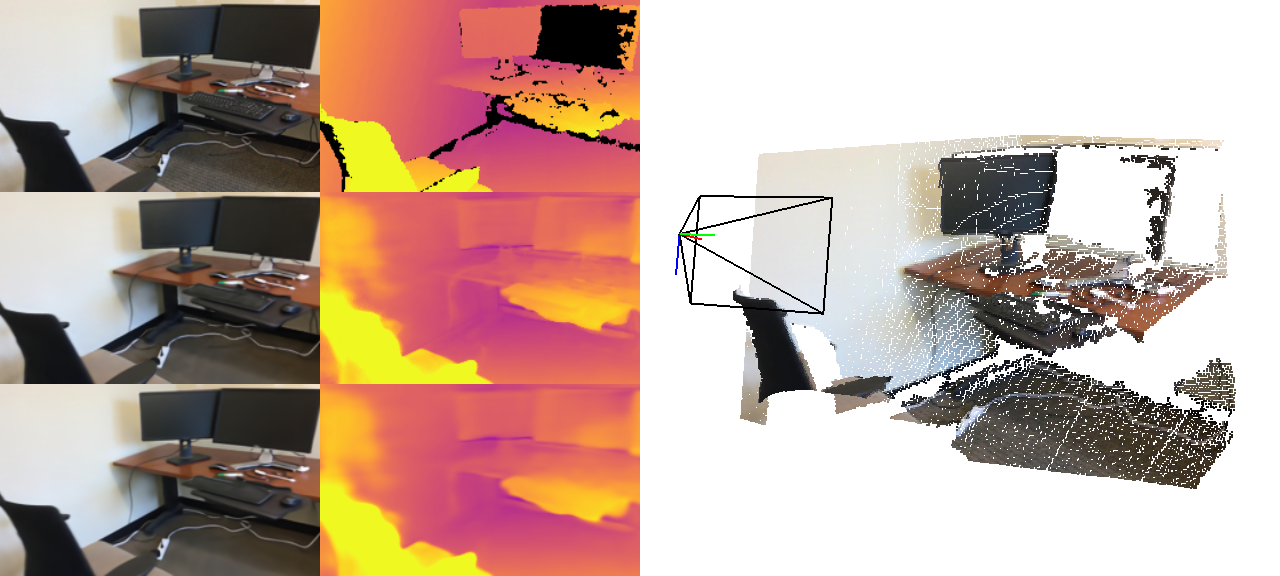} &
  \includegraphics[width=0.3\textwidth,trim=23cm 0cm 0cm 2cm,clip]{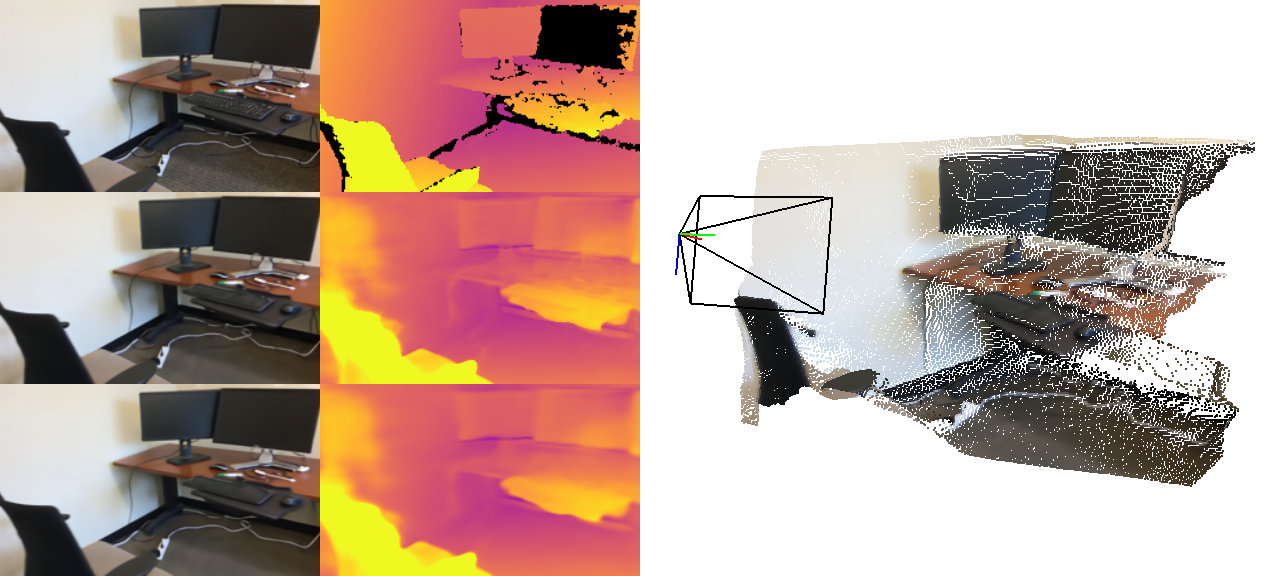} &
  \includegraphics[width=0.3\textwidth,trim=23cm 0cm 0cm 2cm,clip]{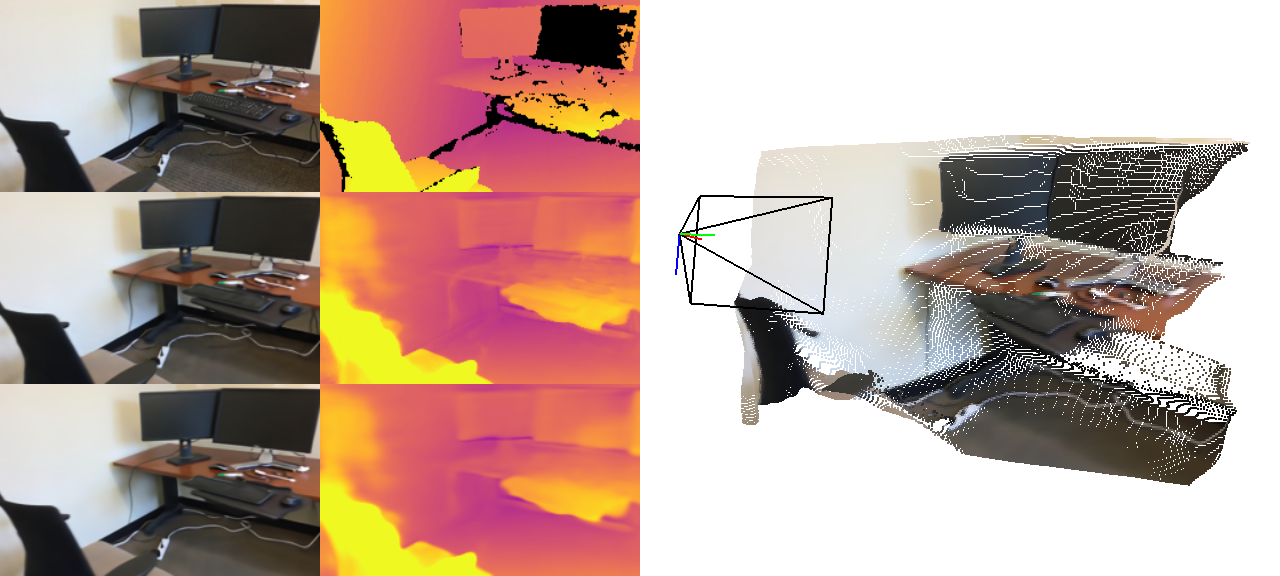}
  \end{tabular}
\caption{
\textbf{Qualitative depth and view synthesis results} from unseen viewpoints, using different \Acronym decoders. The first column shows ground truth point clouds, while the second and third columns show respectively pointclouds generated using radiance field predictions, and depth and light field predictions.
}
\label{fig:pointclouds_supp}
\end{figure*}

Moreover, in \Figure \ref{fig:pointclouds_supp} we show additional point clouds generated from novel viewpoints using different \Acronym decoders, relative to the ground truth point cloud.  Each point cloud is generated by lifting pixel colors to 3D space, using camera intrinsics and depth information. Ground truth point clouds use provided RGB images and depth maps, while predicted pointclouds use estimates for specific decoders (radiance for volumetric renderings, and depth and light fields for single-query renderings). 

\section{Comparison with  Monodepth}
Our multi-view photometric regularization is inspired by the self-supervised loss used in monocular depth estimation. For illustrative purposes, we show in Fig.~\ref{fig:monodepth} a qualitative comparison of depth maps from \Acronym and monodepth2~\cite{monodepth2}, a traditional monocular depth network.
Self-supervised depth estimation requires a large amount of training data to learn accurate predictions, since the multi-view photometric objective is highly ambiguous and has several local failure cases (e.g., reflective surfaces, non-Lambertian objects, textureless areas).  In the indoor setting, where these types of surfaces are common, it is thus highly challenging, and for the example in Fig.~\ref{fig:monodepth} the self-supervised depth network fails to properly capture the observed scene geometry.
In contrast, our method maintains a volumetric representation, which attenuates the effect of the self-supervised photometric loss, thus allowing for the network to more accurately reconstruct non-Lambertian surfaces, using the multi-view photometric loss only as a geometric regularizer that gradually vanishes over time.

\end{document}